\documentclass[conference]{IEEEtran}
\usepackage{fontspec}
\usepackage{polyglossia}
\usepackage{amsmath}
\setmainlanguage{english}
\setotherlanguages{tamil}

\newfontfamily\tamilfont[
    Script=Tamil,
    Scale=1.1
]{NotoSerifTamil-Regular.ttf}
\newfontfamily\hindifont{NotoSerifDevanagari-Regular.ttf}

\newfontfamily\kannadafont[
    Script=Kannada,
    Scale=1.05
]{NotoSerifKannada-Regular.ttf}

\newfontfamily\arabicfont[
    Script=Arabic,
    Scale=1.1
]{Amiri-Regular.ttf}

\newfontfamily\russianfont[
    Script=Cyrillic
]{cmunrm.ttf}

\newfontfamily\greekfont[
    Script=Greek
]{FreeSerif.ttf}
\usepackage{amsmath,hyperref}
\usepackage{graphics,graphicx,color,epsfig,subcaption}
\usepackage{amsfonts}
\usepackage{amsthm,amssymb}
\usepackage{nicefrac}       
\usepackage{xcolor}
\usepackage{multirow}
\usepackage{booktabs}       
\usepackage[table]{xcolor}
\usepackage{colortbl}
\usepackage{tikz}
\usetikzlibrary{automata, positioning, arrows,shapes.multipart,fit}

\usepackage{titlesec}
\usepackage{algorithmic}
\usepackage{algorithm}

\usepackage{latexsym}
\usepackage{amssymb}
\usepackage{inconsolata}
\usepackage[normalem]{ulem}
\usepackage{microtype}

\newfontface\ipafont{CharisSIL-Regular.ttf}
\usepackage{inconsolata}
\usepackage{graphicx}
\usepackage{tikz}
\usetikzlibrary{matrix, arrows.meta, positioning}

\usepackage{multirow}
\usepackage[table]{xcolor}
\definecolor{rank1}{RGB}{230,190,255} 
\definecolor{rank2}{RGB}{200,210,255} 
\definecolor{rank3}{RGB}{140,220,255} 
\definecolor{rank4}{RGB}{200,255,200} 
\definecolor{rank5}{RGB}{255,255,180} 
\definecolor{rank6}{RGB}{255,210,170} 
\definecolor{rank7}{RGB}{255,180,180} 

\usepackage{tikz}
\usetikzlibrary{arrows.meta, positioning, calc}
\usepackage{pgfplots}
\pgfplotsset{compat=1.18}

\newcommand{\reptwo}[1]{\textcolor{black}{#1}}

\title{Breaking the Script Barrier: Enabling Automatic Alignment for PoS-based ASR Error Analysis in Non-Latin Scripts}
\author{
\IEEEauthorblockN{
Prasenjit K Mudi$^{1*}$,
Dahlia Devapriya$^{1*}$,
Sheetal Kalyani$^{1}$
}

\IEEEauthorblockA{
$^{1}$Indian Institute of Technology Madras, India \\
\{ee21d057@smail, ee22d003@smail, skalyani@ee\}.iitm.ac.in \\
$^{*}$Authors contributed equally
}
}
\begin{document}
\maketitle
\begin{abstract}
Automatic Speech Recognition (ASR) systems are commonly evaluated using aggregate metrics such as Word Error Rate (WER), which do not capture the linguistic structure of errors. Fine-grained analysis, such as Part-of-Speech (PoS)-wise error characterization, requires accurate alignment between ASR hypotheses and reference transcriptions. However, existing alignment tools are often unreliable for languages written in non-Latin scripts.
In this work, we address this gap by proposing a robust, automated, language-agnostic alignment mechanism applicable across ASR architectures and across languages written in both Latin and non-Latin scripts. This enables consistent alignment of hypotheses, references, and evaluation sequences, forming the basis for downstream linguistic analysis. Building on this, we employ standard PoS taggers to perform scalable and reproducible PoS-wise error analysis. Notably, we perform alignment and downstream ASR error analysis across three major segmented writing systems, namely, Abugida (Tamil, Hindi, Kannada), Alphabetic (English, Russian, Greek), and Abjad (Arabic). We further demonstrate how such error information can be leveraged during ASR training to improve metrics such as WER.
\end{abstract}

\section{Introduction}
\label{sec:intro}
Automatic Speech Recognition (ASR) has become a widely deployed technology, with applications ranging from voice assistants to large-scale transcription systems. Word Error Rate (WER) remains the predominant evaluation metric \cite{von2025word}, along with alternatives such as Match Error Rate (MER), Word Information Lost (WIL) \cite{morris2004and}, Human Perceived Accuracy (HPA) \cite{mishra2011predicting}, and weighted variants of WER \cite{le2016better}. However, these metrics provide only a coarse-grained view of performance and do not capture the underlying linguistic structure of errors. In particular, they fail to reveal the types of mistakes made in terms of grammatical categories. 
By incorporating Part-of-Speech (PoS)-wise analysis of errors, we can obtain a deeper understanding of how ASR systems handle different grammatical categories.

Performing automated PoS-wise error analysis over substitution (S), insertion (I), and deletion (D) errors requires aligning the ASR hypothesis (HYP), the reference transcription (REF), and an evaluation string (EVAL) that encodes the error types for each utterance. In standard ASR pipelines, tools such as \texttt{sclite} \cite{sclite} are used to generate alignment-based evaluation reports. While \texttt{sclite} provides reliable alignments for English and other Latin-script languages, its alignment quality significantly degrades for languages written in non-Latin scripts, hence making downstream PoS-based error analysis unreliable. Specifically, for languages such as Tamil, Hindi, Kannada, Arabic, Russian and Greek, we observed that the alignment between the REF, HYP, and EVAL fields was often corrupted due to non-uniform character widths and complex characters, which led to inconsistent spacing. In contrast, languages written in the Latin script, suhobxch as English, Spanish, and German, exhibited more uniform spacing, making the default \texttt{sclite} alignment sufficiently reliable. We require an alignment algorithm that is able to adapt to non-uniformity in spacing and complexity of characters. Hence we propose a robust alignment mechanism that enables accurate PoS-wise error analysis across languages and scripts. 

The Needleman-Wunsch algorithm \cite{needleman1970general} was originally developed to align sequences of amino acids to identify similarities and infer evolutionary relationships. It has also been applied to DNA sequence analysis, such as aligning coronavirus genomes to identify mutations \cite{isa2019application}. While widely used in bioinformatics, its application to other domains has been relatively limited. More recently, \cite{bhogale2023effectiveness} employed the Needleman-Wunsch algorithm to align audio and text sequences for constructing the Shrutilipi dataset.

Motivated by its effectiveness in sequence alignment, we propose a character-spacing-aware modified variant of the Needleman–Wunsch algorithm to align predicted and reference text sequences from ASR outputs, along with their corresponding evaluation strings. 
Once the strings are aligned, a PoS tagger is used in order to classify the errors by their parts-of-speech. Note that in standard PoS tagging, only the reference text was assigned PoS labels; therefore, alignment was not required at this stage, since PoS tags are not defined over ASR error strings but over the tokenized reference sequences. However, for PoS-wise error analysis, alignment is crucial and without it one cannot proceed further. 
We employ the spaCy \footnote{\url{https://spacy.io/api/tagger}}, Stanza \cite{qi2020stanza}, AI4Bharat \cite{kakwani2020indicnlp}, and CAMeL \cite{obeid2020camel} PoS taggers in this work.

Based on our PoS-wise error analysis, we integrate this information into the architecture of the vanilla transformer \cite{vaswani2017attention}, obtaining improvement in WER along with grammar-wise improvements. We observe that \reptwo{even} assigning higher weights to tokens belonging to parts of speech that are most prone to errors during the computation of decoder cross-attention, reduces the number of errors in these susceptible tokens, thereby ultimately reducing the WER.

 To the best of our knowledge, this work is the first to (1) provide an automated alignment mechanism to align the outputs of ASR models applicable across languages and architectures based on (a) modified variant of Needleman-Wunsch algorithm, (b) adaptive character spacing, and (c) post-processing modification of evaluation tags, (2) employ PoS taggers once alignment is done, for the automatic inspection of ASR errors, thereby minimizing manual intervention and enabling scalable, reproducible PoS-wise error analysis, and (3) provide an example of how PoS-wise error information can be utilized in training an ASR model. 
We show alignment result for the languages (a) Tamil, (b) Hindi, (c) Kannada, (d) Russian, (e) Greek, \reptwo{(f) Arabic, and (g) English}. Note that all the above languages have widely different written scripts. According to \cite{daniels1996world}, the segmentable languages of the world can be classified into the \reptwo{three writing systems -} Abugida, Alphabetic and Abjad. Languages such as Tamil, Hindi and Kannada, where consonant symbols inherently carry vowels, belong to the Abugida writing system. Languages such as \reptwo{Russian, English,} and Greek, where vowels and consonants are represented as independent symbols, belong to the Alphabetic writing system. Arabic, where writing is primarily consonant-based and vowels are often omitted in standard text, belongs to the Abjad writing system. The alignment by \texttt{sclite} works for languages which have a Latin scipt and belong to the Alphabetic writing system, \reptwo{while our algorithm works for both Latin and non-Latin script languages, breaking the script barrier}. 
\reptwo{We provide details of the alignment mechanism in Section \ref{sec:alignment}, PoS-wise error analysis across  languages in Section \ref{sec:error_analysis} and results on PoS-aware transformer in Section \ref{sec:pos_aware_Transformer}}. 

\section{Related Work}
\textbf{Error analysis in ASR}
Understanding ASR errors has been explored across multiple languages and settings. \reptwo{In \cite{papadopoulou2021benchmarking}, transcript quality is evaluated based on human post-editing effort, while \cite{wirth2022automatic} analyzes German ASR outputs by categorizing errors into context-breaking, phonetic confusions, and annotation inconsistencies across pretrained models. Similarly, \cite{chowdhury2024analyzing} studies Bengali ASR transcripts and shows that readability is not always reflected in WER, with errors often arising from speaker variability and unclear word boundaries.} However, these approaches rely heavily on manual inspection, limiting scalability and reproducibility. In contrast, our work performs automated, linguistically grounded error analysis using PoS information. More recently, \cite{parsons2025match} proposed a multi-tiered alignment framework combining character and word-level alignment with speech-production-based similarity metrics and compound reconciliation for one-to-many and many-to-one mappings, evaluated mainly on European languages with Latin scripts. Unlike this approach, our algorithm does not use phonetic or articulatory information, but instead leverages character-width properties and generalizes across diverse \reptwo{segmented writing} systems including Alphabetic, Abugida, and Abjad.\\
\textbf{PoS integration in model architectures}
Efforts to incorporate linguistic information into model architectures have mainly focused on text-based tasks, where inputs are inherently aligned. For instance, the PL-Transformer \cite{shi2023pl} integrates PoS embeddings for text classification using well-formed textual inputs without requiring sequence alignment. In contrast, ASR outputs require explicit alignment between predicted and reference sequences before any \reptwo{meaningful linguistic analysis can be performed}.

Similarly, prior works in machine translation and sequence modeling leverage PoS information in settings where alignment is already available. In \cite{james2025pos}, word alignment for text-to-text machine translation is improved using PoS-tagged data for Dravidian languages. In \cite{hlaing2022improving}, concatenating PoS embeddings with words improves English--Thai translation, while \cite{wang2021pos} uses PoS tags of preceding words to enhance next-word prediction in video captioning. However, these methods operate on text with implicit one-to-one alignment between tokens and their \reptwo{PoS annotations}.
\section{Existing alignment and proposed approach}\label{sec:alignment}
In this section, we first briefly discuss the default \texttt{sclite} based text alignment and its limitation, followed by our proposed character-spacing aware modified Needleman-Wunsch alignment algorithm.
We use the SPRING\_INX dataset for Tamil, Hindi and Kannada languages\footnote{https://asr.iitm.ac.in/dataset}\cite{r2023springinxmultilingualindianlanguage} and, Common Voice Russian, Greek, Arabic datasets \cite{ardila2020common} for experimentation. For all results illustrated in this Section, we use the transformer model with 12 encoder and 6 decoder blocks for Tamil and Whisper for Russian and Arabic datasets \footnote{\href{https://github.com/Speech-Lab-IITM/espnet/blob/master/egs2/spring_speech/conf/tuning/train_asr_transformer.yaml}{Transformer(Tamil)},\href{https://huggingface.co/antony66/whisper-large-v3-russian}{Whisper(Russian)},\href{https://huggingface.co/mohammed/whisper-small-arabic-cv-11}{Whisper(Arabic)}}. For PoS tagging, we use Stanza integrated with spaCy for Tamil and Hindi, spaCy for Russian, AI4Bharat for Kannada, CAMeL for Arabic and gr-nlp-toolkit for Greek\footnote{\href{https://github.com/explosion/spacy-stanza}{Tamil/Hindi}, \href{https://github.com/explosion/spaCy}{Russian}, \href{https://huggingface.co/ai4bharat/IndicBERTv2-alpha-POS-tagging}{Kannada}, \href{https://github.com/CAMeL-Lab/camel_tools}{Arabic}, \href{https://github.com/nlpaueb/gr-nlp-toolkit}{Greek}}. More details about the PoS taggers are provided in Appendix \ref{ap:pos_tagger}.
\subsection{Default text alignment}
Text alignment is done by using \texttt{sclite} which is a tool for scoring and evaluating ASR systems, developed as part of the NIST SCTK (National Institute of Standards and Technology Scoring Toolkit). It first matches the unique utterance ID from 'HYP' and 'REF' to create REF-HYP pairs. Next, dynamic programming is used to minimize the Levenshtein distance \cite{levenshtein1966binary} between REF and HYP.
The distance metric assigns fixed, non-adaptive costs of 0, 4, 3, and 3 to correct matches, substitutions, insertions, and deletions respectively \cite{sclite}. The final alignment is obtained by globally minimizing the cumulative edit cost over all possible alignments using the \texttt{sclite} framework. Note that
\texttt{sclite} is very popular since it works well most European languages, such as English, Spanish, German, etc., a lot of which have a Latin written script. However, languages which have a non-Latin written script, such as Tamil, Hindi, Russian, Greek, Arabic, etc., suffer from significant alignment issues.
\begin{table}[h!]
\centering
\small
\begin{tabular}{|p{0.8cm}|p{6cm}|}
\hline
REF & humpy ***** dumpy fell downstairs \\
\hline
HYP & humpy don't \hspace{2mm}be  \hspace{4mm}  fell downstairs \\
\hline
Eval & \hspace{10mm}I \hspace{6mm} S \\
\hline
\end{tabular}
\caption{Example illustrating S and I errors in alignment produced by \texttt{sclite} and our proposed algorithm (both produce perfect alignment)}
\label{table:sclite_misalignment_example}
\end{table}
\begin{table}[h]
\centering
\small
\begin{tabular}{|p{2cm}|p{2cm}|p{0.8cm}|p{1cm}|}
\hline
ref\_word & hyp\_word & error & PoS \\
\hline
-        & don't      & I & verb \\
\hline
dumpy    & be         & S & propn \\
\hline
\end{tabular}
\caption{PoS tagging of errors for example in Table~\ref{table:sclite_misalignment_example}.}
\label{table:eng_pos_mapping}
\end{table}
For the sample ASR output shown in Table \ref{table:sclite_misalignment_example}, the ASR model has made one insertion in the second position and one substitution in the third position. Hence, in Eval, \texttt{sclite} aligns `I' to  the first character of the second word, and an `S' with the first character of the third word. When an insertion occurs, i.e., a word is not present in REF but is present in HYP, \texttt{sclite} inserts the character `*' in REF, repeating it for as many characters as the inserted word in HYP, aligned to that word. Similarly, when a deletion occurs, i.e., a word is present in REF but not in HYP, \texttt{sclite} inserts `*' in HYP corresponding to that word in REF. In this example, the second word in HYP was inserted, and hence in REF, five `*' characters are inserted corresponding to the 5-character word inserted in HYP. 
\begin{table*}[h!]
\scriptsize
\centering
\begin{tabular}{|p{0.5cm}|p{14cm}|}
\hline
\multicolumn{2}{|c|}{\textbf{Alignment using sclite}} \\
\hline
REF & {\tamilfont **************** அவளுக்கொரு பட்டு பாவாடை தைக்கணும்ன்னு சொல்லி, ஒரு யோசனை இருக்கு.} \\
\hline
HYP & {\tamilfont அதனால, அவளுக்கு ஒரு பட்பாவோட தைக்கணும்னு சொல்லி, ஒரு யோசனை இருக்கு.} \\
\hline
Eval & I \hspace{4mm} S \hspace{13mm} S \hspace{7mm} S \hspace{22mm} S \hspace{23mm} 
\hspace{48mm} \\
\hline
\multicolumn{2}{|c|}{\textbf{Alignment using proposed algorithm}} \\
\hline
REF & {\tamilfont ********* அவளுக்கொரு பட்டு \hspace{0.5mm}பாவாடை \hspace{2.35mm}தைக்கணும்ன்னு சொல்லி, ஒரு யோசனை இருக்கு.}  \\
\hline
HYP & {\tamilfont அதனால, அவளுக்கு \hspace{5mm}ஒரு \hspace{2mm}பட்பாவோட தைக்கணும்னு \hspace{3mm}சொல்லி, ஒரு யோசனை இருக்கு.}  \\
\hline
Eval & I \hspace{12mm} S \hspace{16mm} S \hspace{5mm} S \hspace{13mm} S \hspace{20mm}  \hspace{50mm}  \\
\hline
\end{tabular}
\vspace{3mm}
\begin{minipage}{0.45\textwidth}
\centering
\footnotesize
\begin{tabular}{|p{2.4cm}|p{2cm}|p{0.3cm}|p{0.6cm}|}
\hline
ref\_word & hyp\_word & tag & PoS \\
\hline
- & {\tamilfont \scriptsize அதனால,} & I & adv \\
\hline
{\tamilfont \scriptsize **************} & {\tamilfont \scriptsize அவளுக்கு} & S & pron \\
\hline
{\tamilfont \scriptsize அவளுக்கொரு} & {\tamilfont \scriptsize ஒரு} & S & pron \\
\hline
{\tamilfont \scriptsize பட்டு} & {\tamilfont \scriptsize பட்பாவோட} & S & noun \\
\hline
{\tamilfont \scriptsize பாவாடை} & {\tamilfont \scriptsize தைக்கணும்னு} & S & noun \\
\hline
\end{tabular}
\vspace{1mm}\\
\textbf{(a) PoS tags after sclite alignment}
\end{minipage}
\hspace{10mm}
\begin{minipage}{0.45\textwidth}
\centering
\footnotesize
\begin{tabular}{|p{2.4cm}|p{2cm}|p{0.3cm}|p{0.6cm}|}
\hline
ref\_word & hyp\_word & tag & PoS \\
\hline
{\tamilfont \footnotesize **************} & {\tamilfont \scriptsize அதனால,} & I & adv \\
\hline
{\tamilfont \scriptsize அவளுக்கொரு} & {\tamilfont \scriptsize அவளுக்கு} & S & pron \\
\hline
{\tamilfont \scriptsize பட்டு} & {\tamilfont \scriptsize ஒரு} & S & noun \\
\hline
{\tamilfont \scriptsize பாவாடை} & {\tamilfont \scriptsize பட்பாவோட} & S & noun \\
\hline
{\tamilfont \scriptsize தைக்கணும்ன்னு} & {\tamilfont \scriptsize தைக்கணும்னு} & S & verb \\
\hline
\end{tabular}
\vspace{1mm}\\
\textbf{(b) PoS tags after proposed alignment}
\end{minipage}
\caption{Comparison between \texttt{sclite} and proposed alignment for a Tamil ASR example 
(``avaḷukkoru paṭṭu pāvāṭai taikkaṇumṉṉu colli, oru yōcaṉai irukku.'':
`so, there is an idea to stitch her a silk skirt')}
\label{table:tamil_alignment_comparison}
\end{table*}
Our objective is to determine the PoS associated with each ASR error. For substitution and deletion errors, we assign the PoS of the REF word, while for insertion errors, we assign the PoS of the inserted HYP word. Thus, if a REF proper noun is recognized as a verb in HYP, the substitution is categorized under “proper noun.” This yields entries \(\mathcal{B} = \{(r,h,o,t)\}\), where \(r,h\) denote REF and HYP words, \(o \in \{=,S,D,I\}\) denotes the alignment operation with `=' denoting a match, and \(t\) is the PoS tag. For English, the aligned REF, HYP, and Eval sequences enable direct mapping of errors to words, after which a PoS tagger is applied to the relevant word. This results in error-PoS mappings (Table~\ref{table:eng_pos_mapping}) for Table~\ref{table:sclite_misalignment_example}.
\begin{table*}[h!]
\footnotesize
\centering
\begin{tabular}{|p{0.5cm}|p{14cm}|}
\hline
\multicolumn{2}{|c|}{\textbf{Alignment using sclite}} \\
\hline
REF & {\russianfont Латинская Америка — это регион с наибольшим в мире     неравенством.} \\
\hline
HYP & {\russianfont Латинская Америка — это регион с наибольшим ** мирьем неравенства.} \\
\hline
Eval & \hspace{120mm} D \hspace{2mm} S \hspace{6mm} S \\
\hline
\multicolumn{2}{|c|}{\textbf{Alignment using proposed algorithm}} \\
\hline
REF & {\russianfont Латинская  Америка  —  это  регион  с  наибольшим  в} \hspace{1mm} {\russianfont мире} \hspace{3mm} {\russianfont неравенством.} \\
\hline
HYP & {\russianfont Латинская  Америка  —  это  регион  с  наибольшим  **  мирьем  неравенства.} \\
\hline
Eval & \hspace{71mm} D \hspace{0.5mm} S \hspace{8.5mm} S \\
\hline
\end{tabular}
\vspace{3mm}
\begin{minipage}{0.45\textwidth}
\centering
\footnotesize
\begin{tabular}{|p{2.4cm}|p{2cm}|p{0.3cm}|p{0.6cm}|}
\hline
ref\_word & hyp\_word & tag & PoS \\
\hline
{\russianfont Латинская} & {\russianfont -} & D & adj \\
\hline
{\russianfont Америка} & {\russianfont Латинская} & S & propn \\
\hline
{\russianfont -} & {\russianfont Америка} & S & punct \\
\hline
\end{tabular}
\vspace{1mm}\\
\textbf{(a) PoS tags after sclite alignment}
\end{minipage}
\hspace{10mm}
\begin{minipage}{0.45\textwidth}
\centering
\footnotesize
\begin{tabular}{|p{2.4cm}|p{2cm}|p{0.3cm}|p{0.6cm}|}
\hline
ref\_word & hyp\_word & tag & PoS \\
\hline
{\russianfont в} & {\russianfont **} & D & adp \\
\hline
{\russianfont мире} & {\russianfont мирьем} & S & noun \\
\hline
{\russianfont неравенством.} & {\russianfont неравенства.} & S & noun \\
\hline
\end{tabular}
\vspace{1mm}\\
\textbf{(b) PoS tags after proposed alignment}
\end{minipage}
\caption{Comparison between \texttt{sclite} and proposed alignment for Russian ASR 
(``Latinskaja Amerika — èto region s naibolʹšim v mire     neravenstvom.'':
`Latin America is the region with the greatest inequality in the world')}
\label{table:russian_alignment_comparison}
\end{table*}
\begin{table}[h!]
\footnotesize
\centering
\begin{tabular}{|p{0.6cm}|p{6cm}|}
\hline
\multicolumn{2}{|c|}{\textbf{Alignment using sclite}} \\
\hline
REF & {\arabicfont  من  سيكذب  لاجلك  ،   سيكذب  عليك. } \\
\hline
HYP & \hspace{1mm}{\arabicfont سيكذب  عليك} ** {\arabicfont ما سيكذب  لاجلك} \\
\hline
Eval & \hspace{0mm} S \hspace{39mm} D \hspace{10mm} S \\
\hline
\multicolumn{2}{|c|}{\textbf{Alignment using proposed algorithm}} \\
\hline
REF & {\arabicfont من سيكذب لأجلك ، سيكذب عليك.} \\
\hline
HYP & {\arabicfont ما سيكذب لأجلك} **{\arabicfont  سيكذب عليك} \\
\hline
Eval & \hspace{0mm} S \hspace{16mm} D \hspace{10mm} S\\
\hline
\end{tabular}
\vspace{3mm}
\centering
\begin{tabular}{|p{1.8cm}|p{2cm}|p{0.3cm}|p{1.2cm}|}
\hline
ref\_word & hyp\_word & tag & PoS \\
\hline
{\arabicfont ما} & {\arabicfont من} & S & prep \\
\hline
{\arabicfont سيكذب} & {\arabicfont -} & D & verb \\
\hline
{\arabicfont سيكذب} & {\arabicfont لأجلك} & S & noun \\
\hline
\end{tabular}
\vspace{1mm}
\textbf{(a)}
\hspace{3mm}
\centering
\begin{tabular}{|p{1.8cm}|p{2cm}|p{0.3cm}|p{1.2cm}|}
\hline
ref\_word & hyp\_word & tag & PoS \\
\hline
{\arabicfont ما} & {\arabicfont من} & S & prep \\
\hline
{\arabicfont ،} & **** & D & punc \\
\hline
{\arabicfont عليك} & {\arabicfont .عليك} & S & pron \\
\hline
\end{tabular}
\vspace{1mm}
\textbf{(b)}
\caption{Arabic ASR alignment using \texttt{sclite} (a) and  proposed method (b)
(``mana sayakadḧaba laajalaka , sayakadḧaba älayaka.":
`whoever lies for you will lie to you')}
\label{table:arabic_alignment_comparison}
\end{table}

\reptwo{However, for several non-Latin script languages such as Tamil, the alignments produced by \texttt{sclite} break after insertions or deletions, causing REF, HYP, and Eval entries to shift out of correspondence, as illustrated in Table ~\ref{table:tamil_alignment_comparison}\footnote{Transliterators used: Indic languages (\href{https://www.aksharamukha.com/converter}{Aksharamukha}) and other languages (\href{https://www.lexilogos.com/keyboard/}{Lexilogos}).}. Although the error types (S, I, D) are often identified correctly, the associated words are no longer properly aligned. Similarly, for Russian (Table ~\ref{table:russian_alignment_comparison}), a deletion causes subsequent words to shift, incorrectly aligning tokens such as {\russianfont Латинская} with a deletion and {\russianfont Америка} with {\russianfont Латинская}, thereby propagating erroneous PoS mappings. For Arabic (Table ~\ref{table:arabic_alignment_comparison}), the deletion of a punctuation symbol disrupts the remaining alignment, leading to incorrect pairings such as {\arabicfont سيكذب} with {\arabicfont لأجلك}. Such misalignments make downstream linguistic analysis unreliable, motivating the need for the proposed alignment correction algorithm. Similar alignment failures for Kannada, Greek and Hindi are shown in Tables ~\ref{table:kannada_alignment_comparison}, \ref{table:greek_alignment_comparison} and \ref{table:hindi_alignment_comparison} (Appendix \ref{ap:illustration}).} Due to the incorrect alignment produced by \texttt{sclite}, it becomes ambiguous which Eval tag corresponds to which REF-HYP word pair, leading to erroneous PoS mappings when a PoS tagger is applied directly. As illustrated in Table \ref{table:tamil_alignment_comparison}, (a), Table \ref{table:russian_alignment_comparison}, (a), and Table \ref{table:arabic_alignment_comparison}, (a), the HYP sentence becomes left or right-shifted relative to the REF sentence, making the resulting PoS analysis unreliable. Consequently, automated linguistic error analysis becomes infeasible. To address this, we propose a character-spacing aware modified Needleman-Wunsch based alignment mechanismn.

\subsection{Needleman-Wunsch based alignment}

Consider two sequences; the reference sequence in REF,
\(
R = (r_1, r_2, \dots, r_N)
\)
and the hypothesis sequence in HYP,
\(
H = (h_1, h_2, \dots, h_M)
\) where each $r_i$ corresponds to the $i^{th}$ word in REF and each $h_i$ corresponds to the $i^{th}$ word in HYP.
Let $c_{sub}$, $c_{ins}$, and $c_{del}$ denote the costs assigned to S, I and D errors respectively.
We define a dynamic programming matrix \(D \in \mathbb{R}^{(N+1)\times(M+1)}\), where \(D(i,j)\) represents the alignment cost between the prefixes \(r_{i}\) and \(h_{j}\).
The alignment is obtained by minimizing a weighted Levenshtein distance using the following recurrence.
\begin{equation}\label{sclite_dist}
    D(i,j) = \min \begin{cases}
D(i-1, j-1) + c_{\text{sub}}(r_i, h_j) \\
D(i-1, j) + c_{\text{del}} \\
D(i, j-1) + c_{\text{ins}}
\end{cases}
\end{equation}
The substitution cost is defined as,
\[
c_{\text{sub}}(r_i, h_j) =
\begin{cases}
0 & \text{if } r_i = h_j \quad \text{\footnotesize (correct match)} \\
1 & \text{if } r_i \neq h_j \quad \text{\footnotesize (substitution)}
\end{cases}
\]

The initialization conditions for the first row and first column are $D(0,0) = 0, \quad D(i,0) = i, \quad D(0,j) = j$. Following this, the formula for filling the rest of the entries of the matrix is given by \eqref{sclite_dist}. Let $\mathcal{A}
= \{(r_k, h_k, o_k)\}_{k=1}^{L}$ denote the set of alignments between the reference and hypothesis sequences, where \(r_k\) and \(h_k\) represent the $k^{th}$ aligned reference and hypothesis tokens respectively, and \(o_k \in \{=, S, D, I\}\) denotes match, substitution, deletion, or insertion operation corresponding to the REF and HYP words $r_k$ and $h_k$ respectively. Here $L$ denotes the number of alignment operations.
The minimum cost alignment is then obtained by minimizing the Needleman-Wunsch edit distance over the set of alignments consistent with the observed evaluation tag sequence \(E\), i.e.,
\begin{equation}\label{nw_objective}
\hat{\mathcal{A}}
=
\arg\min_{\mathcal{A}\in\Omega_E}
D_{\text{NW}}(\mathcal{A}),
\end{equation}
where \(\Omega_E\) denotes the set of admissible alignments satisfying the external S/I/D tag constraints and \(D_{\text{NW}}(\mathcal{A})\) is defined as the cumulative edit cost,
\(
D_{\text{NW}}(\mathcal{A})
=
\sum_{k=1}^{|\mathcal{A}|} c(o_k),
\)
with \(c(o_k)\) representing the corresponding match, substitution, insertion, or deletion cost corresponding to the $k^{th}$ alignment operation $o_k$.
In order to solve the optimization problem in \eqref{nw_objective}, there are two steps to be followed. The first is the forward step comprising the filling of a cost matrix. Next, a backtracking procedure is applied on the matrix in order to obtain the optimal sequence alignment between REF and HYP.
Consider the REF and HYP entries given in Table \ref{table:backtracking_result}. There is one deleted word in HYP, namely the second word, and one inserted word in HYP in the fourth position. Our objective is to align the HYP to REF with minimal cost of (S, D, I).
\begin{table}[h]
\small
    \centering
    \begin{tabular}{|p{0.8cm}|p{4cm}|}
    \hline
    REF & he is \hspace{1mm}going ** home\\
    \hline
    HYP & he ** going to home\\
    \hline
    Eval & \hspace{3mm} D \hspace{8mm} I \\
    \hline
    \end{tabular}
    \caption{An example for illustration of Needleman-Wunsch based alignment}
    \label{table:backtracking_result}
\end{table}\\
\textbf{Matrix construction.}
The cost matrix is constructed using \eqref{sclite_dist} which can be illustrated using the example given in Table \ref{table:backtracking_result}. First, the REF tokens are filled along the rows and the HYP tokens are filled along the columns as shown in Fig. \ref{fig:nw_backtracking}.
Proceeding in this manner, each entry in the row is computed using the same recurrence by considering three possibilities: matching/substitution (diagonal), deletion (up), and insertion (left). This ensures that $D(i,j)$ always stores the minimum cost alignment between the corresponding prefixes.
For the example in Table~\ref{table:backtracking_result}, the fully constructed matrix is shown in Fig. \ref{fig:nw_backtracking}. 
\begin{figure}[h]
\centering
\begin{center}
\begin{tikzpicture}
\matrix (m) [matrix of nodes, nodes={minimum size=8mm, anchor=center},
    row sep=-\pgflinewidth,
    column sep=-\pgflinewidth
] {
    0 & 1 & 2 & 3 & 4 \\
    1 & 0 & 1 & 2 & 3 \\
    2 & 1 & 1 & 2 & 3 \\
    3 & 2 & 1 & 2 & 3 \\
    4 & 3 & 2 & 2 & 2 \\
};
\draw (m-1-1.north west) rectangle (m-5-5.south east);
\foreach \i in {1,...,5} {
    \draw (m-\i-1.south west) -- (m-\i-5.south east);
    \draw (m-1-\i.north east) -- (m-5-\i.south east);
}
\node[left=5mm of m-1-1] {0};
\node[left=5mm of m-2-1] {he};
\node[left=5mm of m-3-1] {is};
\node[left=5mm of m-4-1] {going};
\node[left=5mm of m-5-1] {home};
\node[above=5mm of m-1-1] {0};
\node[above=5mm of m-1-2] {he};
\node[above=5mm of m-1-3] {going};
\node[above=5mm of m-1-4] {to};
\node[above=5mm of m-1-5] {home};
\draw[->, thick, green!70!black, shorten >=2pt, shorten <=2pt]
($(m-5-5.center)+(-0.15,+0.1)$) -- ($(m-4-4.center)+(+0.1,-0.1)$);
\draw[->, thick, green!70!black, shorten >=2pt, shorten <=2pt]
($(m-4-4.center)+(-0.1,0)$) -- ($(m-4-3.center)+(+0.1,0)$);
\draw[->, thick, green!70!black, shorten >=2pt, shorten <=2pt]
($(m-4-3.center)+(-0.15,+0.1)$) -- ($(m-3-2.center)+(+0.1,-0.1)$);
\draw[->, thick, green!70!black, shorten >=2pt, shorten <=2pt]
($(m-3-2.center)+(-0,+0.1)$) -- ($(m-2-2.center)+(-0,-0.1)$); 
\draw[->, thick, green!70!black, shorten >=2pt, shorten <=2pt]
($(m-2-2.center)+(-0.15,+0.1)$) -- ($(m-1-1.center)+(+0.1,-0.1)$);
\end{tikzpicture}
\vspace{5mm}
\end{center}
\caption{Dynamic programming matrix for global alignment between REF and HYP using our proposed algorithm. The highlighted path shows the backtracking corresponding to the minimum edit distance.}
\label{fig:nw_backtracking}
\end{figure}
\\
\textbf{Backtracking.}
The backtracking algorithm used to solve \eqref{nw_objective} can be explained as follows. Starting from the terminal state \((N,M)\), the alignment path is constructed by tracing the predecessor state that produced the minimum cost in \eqref{sclite_dist}. At each step \((i,j)\), one of the following transitions is selected: (1)
\(
(i,j)\rightarrow(i-1,j-1)
\)
corresponding to a match or substitution, (2)
\(
(i,j)\rightarrow(i-1,j)
\)
corresponding to a deletion, or, (3)
\(
(i,j)\rightarrow(i,j-1)
\)
 corresponding to an insertion based on $\min\{D(i-1,j-1),D(i-1,j),D(i,j-1)\}$. The process continues until the initial state \((0,0)\) is reached. In the event of all the neighboring entries having equal cost ($D(i-1,j-1),D(i-1,j),D(i,j-1)$) with each predecessor equally probable, the tie breaking priority is as follows: match(`=') > I > D > S.
The recovered alignment is represented as $\hat{\mathcal{A}} = \{(r_k,h_k,o_k)\}_{k=1}^{L}$ where \(r_k\) and \(h_k\) denote the aligned reference and hypothesis tokens respectively, and \(o_k \in \{=,S,D,I\}\) denotes the corresponding alignment operation.
The detailed illustration of the entire algorithm for the example in Table \ref{table:backtracking_result} is given in Appendix \ref{ap:example}. The optimal alignment is obtained by following the path traced in Fig. \ref{fig:nw_backtracking} leading to the alignment in Table \ref{table:backtracking_result}. 

Summarizing, a diagonal move $(i,j)\rightarrow(i-1,j-1)$ denotes either a correct match ($r_i=h_j$) or a substitution ($r_i\neq h_j$), an upward move $(i,j)\rightarrow(i-1,j)$ denotes a deletion in HYP, and a leftward move $(i,j)\rightarrow(i,j-1)$ denotes an insertion in HYP. Thus, the Needleman–Wunsch algorithm recovers an alignment that identifies deletions and insertions preserving the global structure of the sentence.

\noindent\textbf{Post-processing modification.}
After alignment using the modified Needleman-Wunsch algorithm, we perform a post-processing step.
Specifically, consecutive insertion-deletion pairs (I,D) or deletion-insertion (D,I) pairs are merged into a substitution S, reducing the number of edit operations by one. Such merging avoids artificially inflating the error count by treating a  insertion-deletion pair as two separate errors instead of a single substitution error.
For example, consider the following REF and HYP sequences:
REF: {\tamilfont \footnotesize அவன் வீட்டிற்கு சென்றான்} (``avan vīttirku senrān'':
`he went home');
HYP: {\tamilfont \footnotesize அவன் வீடு சென்றான்.}
Suppose the initial Needleman-Wunsch alignment produces
{
$\mathcal{A}$ = ({\tamilfont \footnotesize அவன்}, {\tamilfont \footnotesize அவன்}, =), 
({\tamilfont \footnotesize வீட்டிற்கு}, -, D), 
(-, {\tamilfont \footnotesize வீடு}, I), 
({\tamilfont \footnotesize சென்றான்}, {\tamilfont \footnotesize சென்றான்}, =)
}.
Here, the alignment contains two non-match operations:
(D,I). They are merged into a single substitution operation S. Hence, the final alignment becomes
$\mathcal{A}_{NW} $ = ({\tamilfont \footnotesize அவன்},{\tamilfont \footnotesize அவன்},=),
({\tamilfont \footnotesize வீட்டிற்கு},{\tamilfont \footnotesize வீடு},S),
({\tamilfont \footnotesize சென்றான்},{\tamilfont \footnotesize சென்றான்},=).

\noindent\textbf{Character width based spacing.}
The misalignment in many non-Latin scripts arises because \texttt{sclite} assumes uniform character widths \footnote{Note that the working of the algorithm is not easy to verify as there does not exist a reliable code base.}, whereas actual rendered widths vary across characters, as seen in Table \ref{table:tamil_alignment_comparison}. For instance, Tamil characters such as {\tamilfont ர்} occupy less space than wider characters like {\tamilfont அ} or {\tamilfont பா}. Since \texttt{sclite} inserts a fixed number of `*' characters for missing tokens without accounting for these width variations, the REF and HYP sequences become misaligned, leading to incorrect Eval tag mapping.
To ensure proper alignment of REF, HYP, and EVAL sequences, especially for ASR outputs of non-Latin scripts, we employ a character width-based spacing mechanism that accounts for the actual display width of each token rather than its raw character count. This is particularly important for Indic and other non-Latin scripts, where display width may differ significantly due to combining characters, vowel modifiers, and other script-specific rendering behavior. Here, \emph{combining characters} are Unicode symbols attached to a base character rather than appearing independently; for example, vowel characters in Tamil and Hindi are often rendered together with consonants as a single unit, which is characteristic of Abugida scripts. Similarly, \emph{vowel modifiers} are marks displayed above, below, before, or after a consonant to alter its pronunciation. Consequently, two tokens with the same character length may still occupy different display widths.

In Hindi, tokens such as
{\hindifont कि} and {\hindifont की}
contain combining vowel markers that alter display width despite comparable character counts. Similarly, Arabic characters are rendered in connected cursive forms and may change shape depending on position within a word; for example,
{\arabicfont كتاب} and {\arabicfont كتــاب}
can occupy different rendered widths despite representing similar character sequences. In Cyrillic scripts like Russian also, character widths are non-uniform, e.g.,
{\russianfont мир} and {\russianfont шум}
contain the same number of characters but occupy different widths.

Hence, standard character-count padding may fail to align REF and HYP tokens correctly because the rendered width of complex non-Latin characters differs from their raw character counts. To address this, we compute token widths using a width-aware rendering function. Let
\(
(r,h,o)
\)
denote an aligned reference token `r', hypothesis token `h', and alignment operation `o'. The corresponding column width is computed as
\[
w = \max\big(\mathrm{width}(r), \mathrm{width}(h), \mathrm{width}(o)\big),
\]
where \(\mathrm{width}(\cdot)\) denotes the rendered token width. Each element is then padded to width
\(
w+\delta,
\)
where \(\delta\) is a small spacing constant for readability and clearer word boundaries; in our implementation, \(\delta=1\). This width-aware spacing mechanism ensures consistent alignment across both Latin and non-Latin scripts, thereby improving readability and downstream linguistic analysis. Using the proposed character-spacing-aware modified Needleman--Wunsch alignment algorithm, the examples in Tables \ref{table:tamil_alignment_comparison}, \ref{table:russian_alignment_comparison}, and \ref{table:arabic_alignment_comparison} now produce correct PoS mappings in their corresponding subfigures (b). Hence, accurate PoS tag mapping is achieved through the proposed alignment method. For example, in Table \ref{table:arabic_alignment_comparison} (b), the deletion is correctly associated with the missing comma ``{\arabicfont ،}", while the substitutions {\arabicfont من} $\rightarrow$ {\arabicfont ما} and {\arabicfont عليك} $\rightarrow$ {\arabicfont .عليك} are aligned correctly. Similar alignment corrections for Hindi, Kannada, and Greek are provided in Appendix \ref{ap:illustration} and an ablation study is provided in Appendix \ref{ap:char_spacing}.
We now proceed with PoS error analysis.
\begin{table*}[t!]
    \centering
    \begin{tabular}{|p{0.75cm}|p{0.8cm}|c|c|c|p{0.8cm}|c|c|c|p{0.75cm}|p{0.8cm}|c|c|c|}
    \hline
    \multirow{2}{*}{\textbf{PoS}} & \multirow{2}{*}{\textbf{count}} & \multicolumn{3}{c|}{\textbf{Tamil \footnotesize(Abugida)}} & \multirow{2}{*}{\textbf{count}} & \multicolumn{3}{c|}{\textbf{Russian \footnotesize(Alphabetic)}} & \multirow{2}{*}{\textbf{PoS}} & \multirow{2}{*}{\textbf{count}} & \multicolumn{3}{c|}{\textbf{Arabic \footnotesize(Abjad)}}\\
    \cline{3-5} \cline{7-9} \cline{12-14}
    & & S & D & I & & S & D & I & & & S & D & I\\
    \hline
    noun  & 18905
    & \cellcolor{rank2}15.4 & \cellcolor{rank2}2.4 & \cellcolor{rank2}3.2 
    & 23934
    & \cellcolor{rank2}4 & \cellcolor{rank2}0.3 & \cellcolor{rank2}0.2
    & noun & 16677
    & \cellcolor{rank2}21.5 & \cellcolor{rank2}0.2 & \cellcolor{rank2}0.1 \\
    \hline
    propn & 15053
    & \cellcolor{rank1}17.5 & \cellcolor{rank1}2.9 & \cellcolor{rank1}3.7 
    & 4158
    & \cellcolor{rank1}6.1 & \cellcolor{rank1}0.5 & \cellcolor{rank1}0.3
    & verb & 10742
    & \cellcolor{rank3}12.7 & \cellcolor{rank3}0.2 & \cellcolor{rank3}0.1 \\
    \hline
    verb  & 10365
    & \cellcolor{rank3}13.7 & \cellcolor{rank3}1.7 & \cellcolor{rank3}2.6 
    & 14020
    & \cellcolor{rank3}3.6 & \cellcolor{rank3}0.6 & \cellcolor{rank3}0.2
    & propn & 8918
    & \cellcolor{rank4} 11.7 & \cellcolor{rank4}0.2 & \cellcolor{rank4}0.1 \\
    \hline
    adj   & 5316
    & \cellcolor{rank4}10.6 & \cellcolor{rank4}2.5 & \cellcolor{rank4}3.1
    & 9490
    & \cellcolor{rank4}3.1 & \cellcolor{rank4}0.2 & \cellcolor{rank4}0.2
    & adj & 2380
    & \cellcolor{rank1}23.9 & \cellcolor{rank1}0.5 & \cellcolor{rank1}0.1 \\
    \hline
    adv   & 5245
    & \cellcolor{rank4}10.7 & \cellcolor{rank4}1.9 & \cellcolor{rank4}3.6
    & 5512
    & \cellcolor{rank6}2.4 & \cellcolor{rank6}0.4 & \cellcolor{rank6}0.2
    & prep & 5378
    & \cellcolor{rank5}4.9 & \cellcolor{rank5}0.1 & \cellcolor{rank5}0.1 \\
    \hline
    pron & 4816
    & \cellcolor{rank3}10.6 & \cellcolor{rank3}3.2 & \cellcolor{rank3}4.2 
    & 8019
    & \cellcolor{rank5}2.9 & \cellcolor{rank5}0.3 & \cellcolor{rank5}0.1
    & sconj & 1680
    & \cellcolor{rank5}4.8 & \cellcolor{rank5}0.2 & \cellcolor{rank5}0.1 \\
    \hline
    \end{tabular}    
    \vspace{0.5em} 
    \begin{tabular}{l l l l l l l}
    \cellcolor{rank1}\phantom{x} Rank 1 &
    \cellcolor{rank2}\phantom{x} Rank 2 &
    \cellcolor{rank3}\phantom{x} Rank 3 &
    \cellcolor{rank4}\phantom{x} Rank 4 &
    \cellcolor{rank5}\phantom{x} Rank 5 &
    \cellcolor{rank6}\phantom{x} Rank 6 &\\
    \end{tabular}
    \caption{Error distributions of Transformer across Tamil, Russian, and Arabic data, ranked by PoS frequency. Each cell reports the percentage of errors for a PoS category, computed as $\frac{\text{\# errors}}{\text{\# occurrences of the PoS}} \times 100$.}
    \label{table:err_analysis_ta_ru_ar}
\end{table*}
\section{Error Analysis}
\label{sec:error_analysis}
After alignment, we use PoS taggers, described as a function defined by $\mathcal{F}: \mathcal{E} \to \mathcal{C}$ where $\mathcal{E}$ is the set of $(S, D, I)$ errors and $\mathcal{C}$ is the set of all Universal PoS tags, namely, \textit{adj}, \textit{adp}, \textit{adv}, \textit{aux}, \textit{cconj}, \textit{det}, \textit{intj}, \textit{noun}, \textit{num}, \textit{part}, \textit{pron}, \textit{propn}, \textit{punct}, \textit{sconj}, \textit{verb}, \textit{x} and, \textit{space}, corresponding to adjective, adposition, adverb, auxiliary, coordinating conjunction, determiner, interjection, noun, numeral, particle, pronoun, proper noun, punctuation, subordinating conjunction, verb, other, and whitespace, respectively. More details about the PoS taggers used is in Appendix \ref{ap:pos_tagger}.
Note that our focus is on emphasizing the importance of PoS-wise error analysis and the utility of the analysis in improving the performance of any model. Hence we focus on the benchmarking models, such as Transformer and Whisper models mentioned in Section \ref{sec:alignment}. The method itself can be applied to any model.
In order to analyze error patterns across three different segmented writing systems, we select Tamil, Russian, and Arabic as representatives of the Abugida, Alphabetic, and Abjad systems, respectively. The $(S, D, I)$ error distributions of the Transformer model on the SPRING INX Tamil dataset and the Whisper model on the Common Voice Russian and Arabic datasets are summarized in Table \ref{table:err_analysis_ta_ru_ar} for six major PoS categories (all PoS categories are provided in Tables \ref{tab:tamil_russian_pos_stats} and \ref{tab:arabic_pos_stats} of Appendix \ref{ap:error_analysis}). The column “count” denotes the total number of words for the corresponding PoS in each dataset. \reptwo{The color-coded cells represent the ranking of PoS categories in descending order of total error percentage $\frac{(S+D+I)*100}{count}$.} The rankings suggest that error patterns may vary across languages.

For Tamil, approximately $20.98\%$ of noun tokens and $24.05\%$ of proper noun tokens are in error. For Russian, approximately $4.56\%$ of noun tokens and $6.88\%$ of proper noun tokens are erroneous. Proper noun errors are likely due to out-of-vocabulary or rare words. Noun errors in Tamil arise from its agglutinative nature, whereas in Russian they stem from its rich inflectional structure. For instance, the Tamil noun {\tamilfont \footnotesize குழந்தை} (“child”) may appear as {\tamilfont \footnotesize குழந்தைக்கு} (“to/for the child”), while the Russian noun {\russianfont книга} (“book”) may appear as {\russianfont книги}, {\russianfont книгу}, or {\russianfont книгой}, where suffix variations alter pronunciation and grammatical meaning. For Arabic, approximately $21.03\%$ of noun tokens and $20.45\%$ of proper noun tokens are erroneous. This can be linked to the morphologically rich nature of Arabic, where words undergo root-pattern transformations and affixation. For example, the noun {\arabicfont كتاب} (“book”) may appear as {\arabicfont الكتاب} (“the book”) or {\arabicfont كتابه} (“his book”), introducing pronunciation and spelling variations. In contrast, verbs, pronouns, adjectives, and adverbs exhibit relatively fewer errors, likely due to their higher frequency and more regular usage patterns, enabling the Transformer to learn more robust representations. \reptwo{The PoS error distribution for the languages Hindi, Kannada, Greek and English are discussed in Appendix \ref{ap:error_analysis}. Further, we show how PoS-wise error distributions shift across the models Transformer, LSTM and Conformer for Tamil ASR in Appendix \ref{ap:error_analysis_tamil}.}
In the next section, we propose a PoS-aware Transformer leveraging the PoS-wise error analysis
to improve performance. 

\section{Proposed PoS aware Transformer}\label{sec:pos_aware_Transformer}
At decoding step \(i\), the decoder attends to encoder outputs to identify acoustic frames relevant for predicting target token. Since the target PoS tag is known during training, we use this information to modulate decoder cross-attention based on empirical PoS-wise error distributions, assigning higher importance to frequently error-prone categories.
The standard scaled dot-product attention is,
\[
\mathrm{Attention}(\mathbf{Q},\mathbf{K},\mathbf{V})
=
\mathrm{softmax}\!\left(
\frac{\mathbf{Q}\mathbf{K}^\top}{\sqrt{d_k}}
\right)\mathbf{V}
\]
where \(\mathbf{Q}, \mathbf{K}, \mathbf{V}\) denote query, key, and value matrices. We introduce a PoS-dependent weight \(w_{\text{POS}(j)}\) for each encoder position \(j\), modifying attention as
\[
\tilde{A}_{ij}
=
\mathrm{softmax}\!\left(
\frac{\mathbf{Q}_i\mathbf{K}_j^\top}{\sqrt{d_k}}
\cdot
w_{\text{POS}(j)}
\right)
\]
Now, the attended output becomes
\(
\mathbf{O}_i = \sum_j \tilde{A}_{ij}\mathbf{V}_j.
\)
Thus, attention scores are reweighted according to syntactic importance inferred from PoS tags. The model is trained using the standard joint CTC-attention objective with label smoothing:
\(
\mathcal{L}_{\text{total}}
=
\lambda_{\text{CTC}}\mathcal{L}_{\text{CTC}}
+
(1-\lambda_{\text{CTC}})\mathcal{L}_{\text{Attn}}.
\)
Here, \(d_k\) denotes the key-vector dimensionality and 
\(\lambda_{\text{CTC}} \in [0,1]\) controls the trade-off.
\\
\reptwo{\noindent \textbf{Experimental setup.}
We apply PoS-aware training for Tamil, Arabic, and Russian using a Transformer model. The PoS weights are chosen from empirical PoS-wise error distributions by assigning larger weights to categories contributing disproportionately to decoding errors. Further details on the experimental setup and PoS weights are provided in Appendix \ref{sec:appendix_poswts}. We compare a baseline model without PoS against the proposed PoS-aware Transformer. The architecture is illustrated in Fig. \ref{fig:decoder_cross_attention_zoom} of Appendix \ref{ap:decoder_pos}.}\\
\reptwo{\noindent \textbf{Performance improvement.}
Tamil ASR results show that PoS aware Transformer consistently reduces WER, character error rate (CER), token error rate (TER) and sentence error rate (S.Err), improving WER/ CER/ TER/ S.Err from 23.3/ 8.7/ 16.2/ 73.9 to 22.0/ 8.3/ 15.2/ 67.1. Similarly, for Arabic, WER/ CER/ TER improve from 41.6/ 17.4/ 22.8 to 41.4/ 16.6/ 21.9, while for Russian they improve from 9.7/ 2.9/ 4.1/ 37.9 to 9.2/ 2.7/ 3.8/ 36.0. This indicates that PoS-aware training contributes to more accurate recognition across both languages. More details of PoS-wise error reduction across PoS categories are in Appendix \ref{ap:decoder_pos}.}
\section{Conclusion and Future Scope}
In this work, we addressed the problem of incorrect REF-HYP alignment in ASR outputs, particularly for non-Latin scripts, which hinders automated PoS-wise error analysis. We showed that \texttt{sclite} fails to align many non-Latin scripts, and proposed a character-spacing aware modified Needleman-Wunsch algorithm that generalizes across both Latin and non-Latin scripts. Using the corrected alignments, we performed automated PoS-wise ASR error analysis and incorporated PoS error statistics into PoS-aware decoding. The promising results indicate that leveraging PoS-wise error analysis for ASR architecture design merits future research.
\section*{Limitations}
The proposed alignment algorithm is primarily designed for segmented writing systems in which words are explicitly separated by whitespace. Consequently, the method is not directly applicable to non-segmented scripts such as Chinese, Japanese, or Thai, where token boundaries are not explicitly marked. Extending the alignment framework to such languages would require an additional word segmentation stage, whose errors may further influence alignment quality and downstream PoS analysis.

Additionally, the PoS-wise error analysis framework depends heavily on the quality of the underlying PoS tagger. In low-resource languages or languages with limited NLP tools, the PoS tagger may produce inaccurate or overly coarse-grained annotations. For instance, in the Kannada experiments, a large number of words were assigned to the \textit{X} category, indicating that the tagger was unable to confidently determine the appropriate syntactic class. Such misclassifications can affect the reliability of the resulting PoS-wise ASR error statistics and PoS-aware transformer training.

Furthermore, the proposed PoS weighting strategy was an early result to show the utility of PoS-wise ASR error analysis. Hence further improvement is possible.

\bibliographystyle{IEEEbib}
\bibliography{output}
\appendix
\begin{table*}[h!]
\footnotesize
\centering
\begin{tabular}{|p{0.5cm}|p{14cm}|}
\hline
\multicolumn{2}{|c|}{\textbf{Alignment using sclite}} \\
\hline
REF & {\kannadafont ಅಹ್ ನಾನ್ಸ ಹುಷಾರ್ ಇದ್ದೇನೆ,} \hspace{23mm}{\kannadafont ಸೌಖ್ಯ ಆಲ್ಲಿಯೆಲ್ಲ ಮನೆಯಲ್ಲಿ ಎಲ್ಲಾ?}\\
\hline
HYP & {\kannadafont ಹಾ} \hspace{5mm}{\kannadafont ನಾನ್ಸ ****************** ಹುಷಾರಿದ್ದೇನೆ ಸೌಖ್ಯ ಅಲ್ಲಿ} \hspace{25mm}{\kannadafont ಎಲ್ಲ} \hspace{6mm}{\kannadafont ಮನೆಯಲ್ಲೆಲ್ಲ.}\\
\hline
Eval & S \hspace{40mm} D \hspace{33mm} S \hspace{48mm} S \hspace{1mm}S\hspace{1mm}S\\
\hline
\multicolumn{2}{|c|}{\textbf{Alignment using proposed algorithm}} \\
\hline
REF & {\kannadafont ಅಹ್ ನಾನ್ಸ}\hspace{2mm}{\kannadafont ಹುಷಾರ್} \hspace{2mm}{\kannadafont ಇದ್ದೇನೆ,}\hspace{9mm}{\kannadafont ಸೌಖ್ಯ}\hspace{2mm}{\kannadafont ಆಲ್ಲಿಯೆಲ್ಲ}\hspace{2mm}{\kannadafont ಮನೆಯಲ್ಲಿ}\hspace{2mm}{\kannadafont ಎಲ್ಲಾ?}\\
\hline
HYP & {\kannadafont ಹಾ} \hspace{2mm}{\kannadafont ನಾನ್ಸ} \hspace{2mm}{\kannadafont ******} \hspace{2mm}{\kannadafont ಹುಷಾರಿದ್ದೇನೆ}\hspace{2mm}{\kannadafont ಸೌಖ್ಯ}\hspace{2mm}{\kannadafont ಅಲ್ಲಿ} \hspace{8mm}{\kannadafont ಎಲ್ಲ} \hspace{9mm}{\kannadafont ಮನೆಯಲ್ಲೆಲ್ಲ.}\\
\hline
Eval & S \hspace{12mm} D \hspace{28mm} S \hspace{6mm} S \hspace{10mm} S \hspace{12mm} S \\
\hline
\end{tabular}
\vspace{3mm}
\begin{minipage}{0.45\textwidth}
\centering
\footnotesize
\begin{tabular}{|p{2.4cm}|p{2cm}|p{0.3cm}|p{0.6cm}|}
\hline
ref\_word & hyp\_word & tag & PoS \\
\hline
{\kannadafont ಅಹ್} & {\kannadafont ಹಾ} & S & intj \\
\hline
{\kannadafont ನಾನ್ಸ} & {\kannadafont -} & D & propn \\
\hline
{\kannadafont ಹುಷಾರ್} & {\kannadafont ನಾನ್ಸ} & S & adj \\
\hline
{\kannadafont ಇದ್ದೇನೆ,} & {\kannadafont ********} & S & X \\
\hline
{\kannadafont ಸೌಖ್ಯ} & {\kannadafont ಹುಷಾರಿದ್ದೇನೆ} & S & noun \\
\hline
{\kannadafont ಆಲ್ಲಿಯೆಲ್ಲ} & {\kannadafont ಸೌಖ್ಯ} & S & adv \\
\hline
\end{tabular}
\vspace{1mm}\\
\textbf{(a) PoS tags after sclite alignment}
\end{minipage}
\hspace{10mm}
\begin{minipage}{0.45\textwidth}
\centering
\footnotesize
\begin{tabular}{|p{2.4cm}|p{2cm}|p{0.3cm}|p{0.6cm}|}
\hline
ref\_word & hyp\_word & tag & PoS \\
\hline
{\kannadafont ಅಹ್} & {\kannadafont ಹಾ} & S & intj \\
\hline
{\kannadafont ಹುಷಾರ್} & {\kannadafont ******} & D & adj \\
\hline
{\kannadafont ಇದ್ದೇನೆ,} & {\kannadafont ಹುಷಾರಿದ್ದೇನೆ} & S & X \\
\hline
{\kannadafont ಆಲ್ಲಿಯೆಲ್ಲ} & {\kannadafont ಅಲ್ಲಿ} & S & adv \\
\hline
{\kannadafont ಮನೆಯಲ್ಲಿ} & {\kannadafont ಎಲ್ಲ} & S & noun \\
\hline
{\kannadafont ಎಲ್ಲಾ?} & {\kannadafont ಮನೆಯಲ್ಲೆಲ್ಲ.} & S & num \\
\hline
\end{tabular}
\vspace{1mm}\\
\textbf{(b) PoS tags after proposed alignment}
\end{minipage}
\caption{Comparison between \texttt{sclite} and proposed alignment for a Kannada ASR example 
(``ah nānsa huṣār iddēne, saukhya ālliyella maneyalli ellā?":
`Ah, I am fine; is everything well at home there?'); 
(a), (b): PoS tags after \texttt{sclite} and proposed alignment respectively.}
\label{table:kannada_alignment_comparison}
\end{table*}
\begin{table*}[h!]
\footnotesize
\centering
\begin{tabular}{|p{0.5cm}|p{14cm}|}
\hline
\multicolumn{2}{|c|}{\textbf{Alignment using sclite}} \\
\hline
REF & {\greekfont ****** Και} \hspace{8mm} {\greekfont πρωτομάστορης ήταν} \hspace{8mm} {\greekfont ο αδελφός μου.} \\
\hline
HYP & {\greekfont και πρωτομά στον} \hspace{24mm} {\greekfont Ρησίταν ο αδερφός μου.} \\
\hline
Eval & I \hspace{8mm}S \hspace{18mm}S \hspace{37mm}S \hspace{25mm}S\\
\hline
\multicolumn{2}{|c|}{\textbf{Alignment using proposed algorithm}} \\
\hline
REF & {\greekfont ******  Και} \hspace{7mm} {\greekfont πρωτομάστορης} {\greekfont ήταν} \hspace{7.5mm} {\greekfont ο} \hspace{2mm}{\greekfont αδελφός} \hspace{2mm}{\greekfont μου.} \\
\hline
HYP & {\greekfont και} \hspace{4mm} {\greekfont πρωτομά  στον} \hspace{15mm} {\greekfont Ρησίταν} \hspace{3.5mm} {\greekfont ο} \hspace{2mm}{\greekfont αδερφός} \hspace{2mm}{\greekfont μου.} \\
\hline
Eval & \hspace{0mm} I \hspace{7mm} S \hspace{10mm} S \hspace{20mm} S \hspace{18mm} S\\
\hline
\end{tabular}
\vspace{3mm}
\begin{minipage}{0.45\textwidth}
\centering
\footnotesize
\begin{tabular}{|p{2.4cm}|p{2cm}|p{0.3cm}|p{0.6cm}|}
\hline
ref\_word & hyp\_word & tag & PoS \\
\hline
- & {\greekfont και} & I & cconj \\
\hline
{\greekfont ******} & {\greekfont πρωτομά} & S & punct \\
\hline
{\greekfont Και} & {\greekfont στον} & S & cconj \\
\hline
{\greekfont πρωτομάστορης} & {\greekfont Ρησίταν} & S & noun \\
\hline
{\greekfont ήταν} & {\greekfont ο} & S & verb \\
\hline
\end{tabular}
\vspace{1mm}\\
(a) PoS tags ater sclite alignment
\end{minipage}
\hspace{10mm}
\begin{minipage}{0.45\textwidth}
\centering
\footnotesize
\begin{tabular}{|p{2.4cm}|p{2cm}|p{0.3cm}|p{0.6cm}|}
\hline
ref\_word & hyp\_word & tag & PoS \\
\hline
***** & {\greekfont και} & I & cconj \\
\hline
{\greekfont και} & {\greekfont πρωτομά} & S & cconj \\
\hline
{\greekfont πρωτομάστορης} & {\greekfont στον} & S & noun \\
\hline
{\greekfont ήταν} & {\greekfont Ρησίταν} & S & verb \\
\hline
{\greekfont αδελφός} & {\greekfont αδερφός} & S & noun \\
\hline
\end{tabular}
\vspace{1mm}\\
(b) PoS tags after proposed alignment
\end{minipage}
\caption{Comparison between \texttt{sclite} and proposed alignment for a Greek ASR example 
(``Kai prōtomástorēs ḗtan o adelphós mou.":
`and my brother was a master craftsman'); 
(a), (b): PoS tags after \texttt{sclite} and proposed alignment respectively.}
\label{table:greek_alignment_comparison}
\end{table*}
\begin{table*}[h!]
\footnotesize
\centering
\begin{tabular}{|p{0.5cm}|p{14cm}|}
\hline
\multicolumn{2}{|c|}{\textbf{Alignment using sclite}} \\
\hline
REF & {\hindifont डर आ रहा है कि ****** हमलोग जाएंगे फिर कुछ होगा की इतना} \\
\hline
HYP & {\hindifont डर आ रहा है कि हम लोग       जाएंगे फिर कुछ होगा की इतना} \\
\hline
Eval & \hspace{46mm} I \hspace{8mm} S \hspace{73mm}\\
\hline
\multicolumn{2}{|c|}{\textbf{Alignment using proposed algorithm}} \\
\hline
REF & {\hindifont डर आ रहा है कि} ****** {\hindifont हमलोग जाएंगे फिर कुछ होगा की इतना}\\
\hline
HYP & {\hindifont डर आ रहा है कि} {\hindifont हम} \hspace{5.5mm} {\hindifont लोग} \hspace{2.5mm} {\hindifont जाएंगे फिर कुछ होगा की इतना}\\
\hline
Eval & \hspace{19mm} I \hspace{8mm} S \\
\hline
\end{tabular}
\vspace{3mm}
\begin{minipage}{0.45\textwidth}
\centering
\footnotesize
\begin{tabular}{|p{2.4cm}|p{2cm}|p{0.3cm}|p{0.6cm}|}
\hline
ref\_word & hyp\_word & tag & PoS \\
\hline
- & {\hindifont डर} & I & noun \\
\hline
{\hindifont डर} & {\hindifont आ} & S & noun \\
\hline
\end{tabular}
\vspace{1mm}\\
(a) PoS tags after sclite alignment
\end{minipage}
\hspace{10mm}
\begin{minipage}{0.45\textwidth}
\centering
\footnotesize
\begin{tabular}{|p{2.4cm}|p{2cm}|p{0.3cm}|p{0.6cm}|}
\hline
ref\_word & hyp\_word & tag & PoS \\
\hline
{\hindifont ******} & {\hindifont हम} & I & pron \\
\hline
{\hindifont हमलोग} & {\hindifont लोग} & S & pron \\
\hline
\end{tabular}
\vspace{1mm}\\
(b) PoS tags after proposed alignment
\end{minipage}
\caption{Comparison between \texttt{sclite} and proposed alignment for a Hindi ASR example 
(``ḍara ā rahā hai ki hamalōga jāēṁgē phira kucha hōgā kī itanā'': `there is fear that if we go, something may happen'); 
(a), (b): PoS tags after \texttt{sclite} and proposed alignment respectively.}
\label{table:hindi_alignment_comparison}
\end{table*}
\section{PoS taggers}\label{ap:pos_tagger}
PoS taggers take as input a sequence of tokens (e.g., a sentence) and predicts a syntactic category label for each token. These predicted labels are treated as the PoS tags of the corresponding words and are subsequently used in our PoS-wise ASR error analysis and PoS-aware decoder training.
SpaCy is an open-source library for advanced natural language processing (NLP) in Python and Cython released under the MIT license\footnote{https://github.com/explosion/spaCy/blob/master/LICENSE}. It supports 70+ languages, among which we have considered Tamil, Hindi, and Kannada. It can be integrated with many NLP models for PoS tagging and has also been used in several Named Entity Recognition (NER) and Information Extraction (IE) tasks in literature\cite{10551087,chantrapornchai2021information,miranda2022multi}.

For Tamil and Hindi, we use Stanza-based PoS taggers integrated through the \texttt{spacy-stanza} wrapper\footnote{\url{https://github.com/explosion/spacy-stanza}}. In this setup, SpaCy provides the tokenization interface and document abstractions, while Stanza performs the underlying linguistic annotation. Stanza is a neural NLP toolkit developed by Stanford NLP and is primarily trained on Universal Dependencies (UD) treebanks \cite{nivre2020universal}. Its PoS tagger is based on bidirectional Long Short-Term Memory (BiLSTM) networks with character-level representations, enabling robust handling of morphologically rich and agglutinative languages. Stanza additionally performs joint morphological feature prediction together with PoS tagging, improving contextual syntactic understanding.

For Kannada, we use the AI4Bharat IndicBERTv2-based PoS tagging model\footnote{\url{https://huggingface.co/ai4bharat/IndicBERTv2-alpha-POS-tagging}}. The model is built upon transformer-based multilingual contextual representations specifically optimized for Indic languages. It leverages IndicBERTv2 embeddings trained on large-scale multilingual Indic corpora and fine-tuned for PoS tagging tasks.

For Russian, we use the PoS tagger available through spaCy\footnote{\url{https://github.com/explosion/spaCy}}, which is trained using UD-style annotations and optimized for efficient syntactic parsing and contextual PoS prediction.

For Arabic, we employ CAMeL Tools\footnote{\url{https://github.com/CAMeL-Lab/camel_tools}}, a comprehensive Arabic NLP toolkit developed by CAMeL Lab. CAMeL Tools provides modules for tokenization, morphological analysis, disambiguation, and PoS tagging specifically tailored for Arabic, which is a morphologically rich language with extensive inflectional and clitic-based variations. The CAMeL PoS tagger utilizes contextual morphological disambiguation together with pretrained language representations and linguistically informed analyzers to generate accurate syntactic labels for Modern Standard Arabic (MSA).

For Greek, we use the \texttt{gr-nlp-toolkit}\footnote{\url{https://github.com/nlpaueb/gr-nlp-toolkit}}, which provides linguistic processing modules specifically designed for Modern Greek. The toolkit supports tokenization, lemmatization, dependency parsing, and PoS tagging, and is trained on Greek corpora annotated under the UD framework. Its contextual tagging approach effectively captures the rich inflectional morphology and syntactic structure of Greek.

Most of the PoS taggers used in this work are trained on UD treebanks \cite{nivre2020universal}, which provide a consistent multilingual annotation framework based on Universal PoS tags (UPoS). The use of UPoS enables cross-lingual consistency in syntactic analysis and facilitates comparative PoS-wise ASR error evaluation across multiple languages with diverse grammatical structures.

 \section{Proposed Needleman-Wunsch based alignment example}\label{ap:example}
 \subsection{Matrix construction}
 Consider the row corresponding to $r_2 = \text{``is''}$ in Fig. ~\ref{fig:nw_backtracking}. This row computes $D(2,j)$ for all $j$ by comparing the prefix $(\text{he, is})$ with increasing prefixes of the hypothesis.

For example, consider the computation of $D(2,2)$, which aligns $(\text{he, is})$ with $(\text{he, going})$. Using \eqref{sclite_dist},
\[
D(2,2) = \min
\begin{cases}
D(1,1) + c_{\text{sub}}(\text{is}, \text{going}) \\
D(1,2) + 1 \\
D(2,1) + 1
\end{cases}
\]
Since $\text{is} \neq \text{going}$, we have $c_{\text{sub}}(\text{is}, \text{going}) = 1$. Substituting the values from the matrix:
\[
D(2,2) = \min
\begin{cases}
0 + 1 \\
1 + 1 \\
1 + 1
\end{cases}
= 1.
\]

Similarly, $D(2,1)$ aligns $(\text{he, is})$ with $(\text{he})$:
\[
\begin{aligned}
D(2,1) &= \min
\begin{cases}
D(1,0) + c_{\text{sub}}(\text{is}, \text{he}) \\
D(1,1) + 1 \\
D(2,0) + 1
\end{cases} \\
&= \min(1+1,\, 0+1,\, 2+1) = 1.
\end{aligned}
\]
which corresponds to deleting ``is''.
\subsection{Backtracking}
Consider the example shown in Fig. \ref{fig:nw_backtracking}. Starting from $(i,j)=(4,4)$, the reference and hypothesis tokens are both \texttt{home}. Since
$
D(4,4)=D(3,3)+0,
$
the current cost could only have been obtained from the diagonal cell through a zero-cost match transition during the matrix filling stage. Hence, we move diagonally to $(3,3)$, appending
$
(\text{home},\text{home},=)
$
to the alignment set $\mathcal{A}$.
At $(3,3)$, the tokens are \texttt{going} and \texttt{to}, which do not match. We now compare the neighboring cells:
$
D(2,3)=2,\qquad D(3,2)=1,\qquad D(2,2)=1.
$
Since the leftward transition produces the minimum valid cost corresponding to an insertion operation,
$
D(3,3)=D(3,2)+1,
$
we move left to $(3,2)$, appending
$
(-,\text{to},I)
$
to $\mathcal{A}$.
Next, at $(3,2)$, the tokens \texttt{going} and \texttt{going} match. Since
$
D(3,2)=D(2,1)+0,
$
the minimum cost must have originated from the diagonal predecessor. Therefore, we move diagonally to $(2,1)$, appending
$
(\text{going},\text{going},=).
$
At $(2,1)$, the tokens \texttt{is} and \texttt{he} do not match. Comparing neighboring cells shows that the top transition satisfies
$
D(2,1)=D(1,1)+1,
$
corresponding to a deletion. Hence, we move upward to $(1,1)$, appending
$
(\text{is},-,D).
$
Finally, at $(1,1)$, the tokens \texttt{he} and \texttt{he} match and satisfy
$
D(1,1)=D(0,0)+0.
$
Thus, we move diagonally to $(0,0)$, appending
$
(\text{he},\text{he},=).
$
The resulting alignment set is therefore
$
\mathcal{A}
=
\{
(\text{home},\text{home},=),
(-,\text{to},I),
(\text{going},\text{going},=),
(\text{is},-,D),
(\text{he},\text{he},=)
\}.
$
Hence the optimal alignment is given by the sequence 
in Table \ref{table:backtracking_result}.

\section{Alignment illustration for Hindi, Kannada, and Greek}\label{ap:illustration}
For all alignments, we set the values of $c_{ins}$ and $c_{del}$ as $1$. We use the Transformer model with 12 encoder and 6 decoder blocks\footnote{\href{https://github.com/espnet/espnet/blob/master/egs2/librispeech/asr1/conf/tuning/train\_asr\_transformer.yaml}{Transformer 12 layer encoder and 6 layer decoder}} for Kannada and Hindi while we use the finetuned Whisper small model for Greek language\footnote{\url{https://huggingface.co/mozilla-ai/whisper-small-el}}.

The datasets and pretrained models used in this work are publicly available under their respective licenses. The Mozilla Common Voice datasets are distributed under the CC0-1.0 license, while the SPRING-INX datasets are released under the CC BY 4.0 license. All Whisper-based pretrained and fine-tuned models used in our experiments are distributed under the Apache-2.0 license. All assets were used in accordance with their respective terms of use and distribution.

\textbf{Kannada.} As illustrated in Table~\ref{table:kannada_alignment_comparison}, the \texttt{sclite} alignment fails to preserve correct word boundaries after the deletion error, leading to a cascade of incorrect substitutions. In the reference sentence, the word {\kannadafont ಹುಷಾರ್} is deleted in the hypothesis; however, \texttt{sclite} incorrectly shifts subsequent alignments and maps unrelated words against each other. Consequently, PoS mappings become unreliable, with words such as {\kannadafont ನಾನ್ಸ} being aligned with a deletion and {\kannadafont ಹುಷಾರ್} being incorrectly mapped to {\kannadafont ನಾನ್ಸ}. This error propagation further causes semantically unrelated substitutions such as {\kannadafont ಸೌಖ್ಯ} $\rightarrow$ {\kannadafont ಹುಷಾರಿದ್ದೇನೆ}. In contrast, the proposed alignment algorithm preserves proper token boundaries after the deletion occurs. The deleted token is correctly associated with {\kannadafont ಹುಷಾರ್}, while subsequent substitutions such as {\kannadafont ಆಲ್ಲಿಯೆಲ್ಲ} $\rightarrow$ {\kannadafont ಅಲ್ಲಿ} are aligned more accurately. As a result, the PoS-wise mappings become considerably more reliable and interpretable for downstream linguistic error analysis.

\textbf{Greek.} Table~\ref{table:greek_alignment_comparison} demonstrates that the \texttt{sclite} alignment produces severe alignment drift after the insertion error at the beginning of the sentence. The inserted word {\greekfont και} causes subsequent words to be incorrectly shifted, leading to erroneous substitutions across the remainder of the utterance. As a result, unrelated mappings such as {\greekfont Και} $\rightarrow$ {\greekfont στον} and {\greekfont ήταν} $\rightarrow$ {\greekfont ο} are produced. This also corrupts the PoS assignments, with several substitutions incorrectly tagged as \textit{punct}. In contrast, the proposed alignment algorithm preserves local word structure despite the insertion. The inserted token is correctly isolated, while meaningful substitutions such as {\greekfont αδελφός} $\rightarrow$ {\greekfont αδερφός} are aligned appropriately. Consequently, the resulting PoS-wise mappings are substantially cleaner and better reflect the true ASR errors present in the hypothesis.

\textbf{Hindi.} As shown in Table~\ref{table:hindi_alignment_comparison}, the \texttt{sclite} alignment struggles to handle the insertion of the split token {\hindifont हम लोग} corresponding to the reference token {\hindifont हमलोग}. Due to this insertion, \texttt{sclite} propagates alignment errors toward the beginning of the utterance, incorrectly aligning unrelated neighboring words. For example, words such as {\hindifont डर} are falsely marked as substitutions even though they are correctly recognized. This results in misleading PoS assignments such as \textit{noun}. In contrast, the proposed alignment algorithm preserves the correct alignments for the unaffected words and localizes the error around the actual mismatch. The insertion corresponding to {\hindifont हम} is correctly identified, while the substitution {\hindifont हमलोग} $\rightarrow$ {\hindifont लोग} is aligned properly. Therefore, the proposed method yields substantially more accurate PoS-wise error mappings and avoids the cascading alignment failures observed in the \texttt{sclite} output.

\section{Ablation Study on adaptive character spacing}\label{ap:char_spacing}

To evaluate the contribution of adaptive character spacing, we perform an ablation study by removing the spacing correction component and applying only the standard Needleman--Wunsch (NW) alignment algorithm. Although pure NW alignment can still identify approximate substitution regions, the absence of adaptive spacing causes severe alignment drift when insertions or deletions occur between tokens of different lengths.

\begin{table*}[h!]
\centering
\scriptsize

\begin{tabular}{|p{0.7cm}|p{14cm}|}
\hline
\multicolumn{2}{|c|}{\textbf{Pure Needleman--Wunsch Alignment}} \\
\hline

REF &
{\tamilfont **************** அவளுக்கொரு பட்டு பாவாடை தைக்கணும்ன்னு சொல்லி, ஒரு யோசனை இருக்கு.}  \\
\hline

HYP &
{\tamilfont அதனால,\hspace{12mm} அவளுக்கு \hspace{5mm}ஒரு \hspace{3mm}பட்பாவோட தைக்கணும்னு சொல்லி, ஒரு யோசனை இருக்கு.}  \\
\hline

Eval &
I \hspace{21mm}
S \hspace{18mm}
S \hspace{6mm}
S \hspace{6mm}
S \hspace{65mm} \\
\hline

\multicolumn{2}{|c|}{\textbf{Proposed Alignment with Adaptive Character Spacing}} \\
\hline

REF &
{\tamilfont ********* அவளுக்கொரு பட்டு \hspace{0.5mm}பாவாடை \hspace{2.35mm}தைக்கணும்ன்னு சொல்லி, ஒரு யோசனை இருக்கு.}  \\
\hline

HYP &
{\tamilfont அதனால, அவளுக்கு \hspace{5mm}ஒரு \hspace{2mm}பட்பாவோட தைக்கணும்னு \hspace{3mm}சொல்லி, ஒரு யோசனை இருக்கு.}  \\
\hline

Eval &
I \hspace{12mm}
S \hspace{16mm}
S \hspace{5mm}
S \hspace{13mm}
S \hspace{20mm}
 \hspace{48mm}\\
\hline

\end{tabular}

\vspace{5mm}


\begin{minipage}{0.47\textwidth}
\centering
\footnotesize

\begin{tabular}{|p{2.7cm}|p{1.85cm}|p{0.3cm}|p{0.55cm}|}
\hline
\multicolumn{4}{|c|}{\textbf{Pure Needleman--Wunsch}} \\
\hline
ref\_word & hyp\_word & tag & PoS \\
\hline

{\tamilfont ***************} &
{\tamilfont \footnotesize அதனால} &
I &
adv \\
\hline

{\tamilfont \footnotesize அவளுக்கொரு} &
{\tamilfont \footnotesize அவளுக்கு} &
S &
pron \\
\hline

{\tamilfont \footnotesize பட்டு} &
{\tamilfont \footnotesize ஒரு} &
S &
noun \\
\hline

{\tamilfont \footnotesize பாவாடை} &
{\tamilfont \footnotesize பட்பாவோட} &
S &
noun \\
\hline

{\tamilfont \footnotesize தைக்கணும்ன்னு} &
None &
S &
verb \\
\hline



\end{tabular}

\vspace{1mm}
\textbf{(a) PoS mapping after pure NW alignment}

\end{minipage}
\hspace{8mm}
\begin{minipage}{0.47\textwidth}
\centering
\footnotesize

\begin{tabular}{|p{2.7cm}|p{2.25cm}|p{0.25cm}|p{0.55cm}|}
\hline
\multicolumn{4}{|c|}{\textbf{Proposed Alignment}} \\
\hline
ref\_word & hyp\_word & tag & PoS \\
\hline

{\tamilfont **************} &
{\tamilfont \footnotesize அதனால,} &
I &
adv \\
\hline

{\tamilfont \footnotesize அவளுக்கொரு} &
{\tamilfont \footnotesize அவளுக்கு} &
S &
pron \\
\hline

{\tamilfont \footnotesize பட்டு} &
{\tamilfont \footnotesize ஒரு} &
S &
noun \\
\hline

{\tamilfont \footnotesize பாவாடை} &
{\tamilfont \footnotesize பட்பாவோட} &
S &
noun \\
\hline

{\tamilfont \footnotesize தைக்கணும்ன்னு} &
{\tamilfont \footnotesize தைக்கணும்னு} &
S &
verb \\
\hline


\end{tabular}

\vspace{1mm}
\textbf{(b) PoS mapping after proposed alignment}

\end{minipage}

\caption{Comparison between pure Needleman-Wunsch alignment and the proposed character-spacing-aware modified Needleman-Wunsch alignment for a Tamil ASR example. Without adaptive spacing, alignment drift causes several substitutions to collapse into incorrect or missing mappings, leading to shifted PoS assignments and multiple \texttt{None}/\texttt{UNK} entries. The proposed method preserves token correspondence and enables reliable PoS-aware ASR error analysis.}

\label{tab:visual_alignment_comparison}
\end{table*}
Table~\ref{tab:visual_alignment_comparison} illustrates this issue using a Tamil ASR example. In the pure NW alignment, a large placeholder token ({\tamilfont ****************}) is inserted to represent the initial insertion {\tamilfont அதனால,}. Since the placeholder length is substantially larger than the actual inserted token, the entire hypothesis sequence becomes right-shifted relative to the reference sequence. This shift propagates through the remaining alignment, causing subsequent words to be matched at incorrect character positions.

As a consequence, downstream PoS mapping becomes corrupted. Several later substitutions collapse entirely into invalid mappings such as \texttt{None}. Thus, even though some substitution locations are approximately identified, the resulting PoS analysis becomes unreliable because the token boundaries are no longer synchronized between the reference and hypothesis sequences.


\begin{table*}[t]
\centering
\footnotesize
\setlength{\tabcolsep}{4pt}

\begin{tabular}{|c|r|r|r|r|c||c|r|r|r|r|c|}
\hline

\multicolumn{6}{|c||}{\textbf{Tamil}} &
\multicolumn{6}{c|}{\textbf{Russian}} \\

\hline

\textbf{POS} & \textbf{Cnt} & \textbf{D} & \textbf{S} & \textbf{I} & \textbf{Tot (\%)} &
\textbf{POS} & \textbf{Cnt} & \textbf{D} & \textbf{S} & \textbf{I} & \textbf{Tot (\%)} \\

\hline

noun  & 18905 & 451 & 2911 & 605 & 20.98 &
propn & 4158  & 22 & 252 & 12 & 6.88 \\

propn & 15053 & 433 & 2630 & 557 & 24.05 &
noun   & 23934 & 74 & 964 & 53 & 4.56 \\

verb  & 10365 & 182 & 1423 & 270 & 18.09 &
part   & 2962 & 42 & 67 & 20 & 4.36 \\

pron  & 4816 & 153 & 508 & 201 & 17.90 &
cconj & 3341 & 49 & 60 & 32 & 4.22 \\

adj & 5316 & 132 & 562 & 163 & 16.12 &
num & 724 & 9 & 18 & 3 & 4.14 \\

adv & 5245 & 102 & 563 & 191 & 16.32 &
verb & 14020 & 36 & 499 & 22 & 3.97 \\

part & 3863 & 87 & 338 & 103 & 13.67 &
adj & 9490 & 22 & 294 & 16 & 3.50 \\

num & 1748 & 74 & 366 & 66 & 28.95 &
pron & 8019 & 26 & 233 & 9 & 3.34 \\

aux & 2789 & 37 & 284 & 73 & 14.13 &
adv & 5512 & 22 & 135 & 10 & 3.03 \\

adp & 1432 & 25 & 233 & 33 & 20.32 &
det & 4110 & 9 & 93 & 6 & 2.63 \\

det & 2131 & 29 & 153 & 82 & 12.39 &
adp & 8112 & 72 & 80 & 45 & 2.43 \\

cconj & 188 & 1 & 3 & 2 & 3.19 &
sconj & 1428 & 3 & 29 & 1 & 2.31 \\

-- & -- & -- & -- & -- & -- &
aux & 1152 & 6 & 14 & 4 & 2.08 \\

\hline
\end{tabular}
\caption{PoS-wise ASR error statistics for Tamil and Russian datasets. D, S, and I denote deletion, substitution, and insertion errors, respectively. Total (\%) denotes the percentage of erroneous tokens within each PoS category.}
\label{tab:tamil_russian_pos_stats}
\end{table*}
In contrast, the proposed alignment algorithm dynamically adjusts character spacing to preserve token boundary consistency after insertions and deletions. As shown in Table~\ref{tab:visual_alignment_comparison}, this prevents cascading right-shift errors and enables semantically meaningful alignments. Words such as {\tamilfont பாவாடை} and {\tamilfont தைக்கணும்ன்னு} are correctly aligned with their corresponding hypothesis tokens, resulting in accurate noun- and verb-level PoS assignments. This demonstrates that adaptive character spacing is essential not only for improving visual alignment quality, but also for obtaining reliable downstream PoS-aware ASR error analysis.


\begin{table}[t]
\centering
\footnotesize
\setlength{\tabcolsep}{4pt}

\begin{tabular}{|c|r|r|r|r|c|}
\hline

\textbf{POS} & \textbf{Cnt} & \textbf{D} & \textbf{S} & \textbf{I} & \textbf{Tot (\%)} \\

\hline

foreign        & 31    & 3  & 42   & 0 & 145.16 \\
abbrev         & 51    & 0  & 16   & 0 & 31.37 \\
part\_det      & 18    & 0  & 5    & 0 & 27.78 \\
adj            & 2380  & 12 & 570  & 1 & 24.50 \\
noun           & 16677 & 39 & 3581 & 7 & 21.75 \\
verb           & 10742 & 18 & 2112 & 6 & 19.88 \\
interj         & 20    & 0  & 3    & 0 & 15.00 \\
noun\_prop     & 8918  & 15 & 1050 & 6 & 12.01 \\
adv            & 336   & 2  & 32   & 1 & 10.42 \\
pron\_interrog & 110   & 0  & 11   & 0 & 10.00 \\
conj           & 903   & 6  & 74   & 1 & 8.97 \\
part           & 201   & 0  & 17   & 0 & 8.46 \\
part\_focus    & 39    & 0  & 3    & 0 & 7.69 \\
pron           & 903   & 2  & 57   & 1 & 6.64 \\
pron\_dem      & 822   & 0  & 53   & 1 & 6.57 \\
pron\_rel      & 1177  & 1  & 64   & 1 & 5.61 \\
noun\_quant    & 111   & 0  & 4    & 2 & 5.41 \\
part\_neg      & 1229  & 4  & 61   & 0 & 5.29 \\
conj\_sub      & 1680  & 4  & 81   & 2 & 5.18 \\
prep           & 5378  & 7  & 266  & 3 & 5.13 \\
part\_verb     & 358   & 0  & 18   & 0 & 5.03 \\
verb\_pseudo   & 134   & 0  & 6    & 0 & 4.48 \\
part\_voc      & 479   & 1  & 20   & 0 & 4.38 \\
adv\_rel       & 79    & 0  & 3    & 0 & 3.80 \\
adv\_interrog  & 117   & 0  & 3    & 0 & 2.56 \\
part\_interrog & 355   & 0  & 7    & 0 & 1.97 \\

\hline
\end{tabular}

\caption{PoS-wise ASR error statistics for the Arabic dataset. D, S, and I denote deletion, substitution, and insertion errors, respectively. Total (\%) denotes the percentage of erroneous tokens within each PoS category.}
\label{tab:arabic_pos_stats}
\end{table}

\section{PoS-wise error analysis}\label{ap:error_analysis}
In all the tabulated results in this section, the \textbf{PoS} column denotes the part-of-speech category assigned to the reference or inserted hypothesis token, while \textbf{Count} indicates the total number of occurrences of that PoS category in the dataset. The columns \textbf{D}, \textbf{S}, and \textbf{I} represent the number of deletion, substitution, and insertion errors respectively associated with that PoS category. The \textbf{Total (\%)} column reports the overall number of errors for that PoS category, computed as D+S+I, along with the corresponding error percentage relative to its total count. Note that, for all datasets, the REF and HYP sentences are normalized by removing punctuation marks. 

Tables \ref{tab:tamil_russian_pos_stats} and \ref{tab:arabic_pos_stats} provide the detailed PoS-wise error analysis for Tamil, Russian and Arabic datasets of which a sub-table is given in Table \ref{table:err_analysis_ta_ru_ar}. It shows clear cross-lingual variation in PoS-wise ASR error distributions. Across Tamil and Arabic, content-bearing categories such as nouns, proper nouns, verbs, and adjectives exhibit substantially higher error rates, dominated primarily by substitution errors. Tamil proper nouns and numerals are particularly error-prone, likely due to vocabulary sparsity and morphological complexity. In contrast, Russian demonstrates comparatively lower and more uniform error rates across PoS categories, reflecting the relative regularity of alphabetic orthography. For Arabic, high substitution rates in nominal and verbal categories indicate the impact of orthographic ambiguity and rich morphology. Overall, these findings highlight the importance of incorporating fine-grained PoS-aware linguistic information into ASR modeling and decoding.
\begin{table}[h]
\centering
\footnotesize
\setlength{\tabcolsep}{4pt}

\begin{tabular}{|c|c|r|r|r|c|}
\hline
\textbf{POS} & \textbf{Count} & \textbf{D} & \textbf{S} & \textbf{I} & \textbf{Total (\%)} \\
\hline
noun  & 16557 & 320 & 2461 & 518 & 3299 (19.93) \\
pron  & 12647 & 396 & 1493 & 586 & 2475 (19.57) \\
verb  & 11885 & 306 & 1464 & 412 & 2182 (18.36) \\
aux   & 10538 & 279 & 1162 & 407 & 1848 (17.54) \\
propn & 4578  & 206 & 1289 & 276 & 1771 (38.69) \\
adp   & 10502 & 184 & 890  & 340 & 1414 (13.46) \\
adj   & 4635  & 104 & 736  & 144 & 984 (21.23) \\
part  & 5132  & 195 & 424  & 226 & 845 (16.47) \\
det   & 3693  & 89  & 401  & 149 & 639 (17.30) \\
sconj & 2788  & 82  & 310  & 125 & 517 (18.54) \\
adv   & 1993  & 40  & 218  & 100 & 358 (17.96) \\
num   & 1754  & 57  & 136  & 107 & 300 (17.10) \\
cconj & 2113  & 67  & 148  & 79  & 294 (13.91) \\
intj  & 409   & 21  & 137  & 17  & 175 (42.79) \\
x     & 91    & 4   & 19   & 0   & 23 (25.27) \\
\hline
\end{tabular}
\caption{PoS-wise error statistics for Hindi.}
\label{tab:hindi_all_pos}
\end{table}

\textbf{Hindi.} For Hindi in Table \ref{tab:hindi_all_pos}, a similar trend is observed where nouns and pronouns account for the highest error counts, followed by verbs. In particular, nouns contribute the largest number of substitution errors, while pronouns exhibit the highest deletion and insertion errors among all PoS categories. Unlike Tamil, Hindi is less morphologically agglutinative but exhibits rich inflectional variations (e.g., gender, number, and case), which can still lead to recognition ambiguities. Pronouns, despite being frequent, show a relatively high error rate, possibly due to their phonetic similarity (e.g., {\hindifont “वह”}, {\hindifont “यह”}) and contextual dependence. Verbs also contribute significantly to errors due to variations in tense, aspect, and agreement markers. Overall, while the Transformer model captures frequent functional categories reasonably well, lexical categories such as nouns remain the primary source of errors in both languages.

\begin{table}[h]
\centering
\footnotesize
\setlength{\tabcolsep}{4pt}
\begin{tabular}{|c|c|r|r|r|c|}
\hline
\textbf{POS} & \textbf{Count} & \textbf{D} & \textbf{S} & \textbf{I} & \textbf{Total (\%)} \\
\hline
x      & 6392 & 402 & 2997 & 351 & 3750 (58.67) \\
adj    & 6641 & 408 & 2505 & 399 & 3312 (49.87) \\
sym    & 5563 & 358 & 1961 & 421 & 2740 (49.25) \\
propn  & 4193 & 318 & 1761 & 341 & 2420 (57.72) \\
verb   & 3226 & 411 & 1262 & 275 & 1948 (60.38) \\
pron   & 1029 & 150 & 774  & 107 & 1031 (100.19) \\
intj   & 1702 & 98  & 653  & 111 & 862 (50.65) \\
num    & 1332 & 132 & 487  & 127 & 746 (56.01) \\
adp    & 741  & 39  & 401  & 16  & 456 (61.54) \\
part   & 640  & 38  & 233  & 43  & 314 (49.06) \\
aux    & 540  & 32  & 219  & 45  & 296 (54.81) \\
noun   & 232  & 15  & 134  & 8   & 157 (67.67) \\
det    & 335  & 16  & 107  & 24  & 147 (43.88) \\
adv    & 122  & 9   & 31   & 11  & 51 (41.80) \\
\hline
\end{tabular}
\caption{PoS-wise error statistics for Kannada.}
\label{tab:kannada_all_pos}
\end{table}
\textbf{Kannada.} The error distribution on the Kannada dataset in Table \ref{tab:kannada_all_pos} reveals a strong dominance of substitution errors (S) across almost all parts of speech similar to Tamil, Hindi. This is particularly evident in high-frequency categories such as adj, sym, propn, and verb, where substitutions account for the majority of total errors, suggesting confusion among acoustically or morphologically similar words. Notably, pron exhibits high total error rate (~100\%), driven largely by substitutions, which reflects the short, highly context-dependent nature of pronouns in Kannada. Punct, noun, and adp also show high overall error rates (>60\%), pointing to challenges in modeling syntactic boundaries and function-word behavior. In contrast, relatively lower error rates for adv, det, and part suggest that these categories are either less frequent, more acoustically distinct, or more predictable given context. Deletion errors (D) are generally smaller compared to substitutions but become more noticeable in verb and pron, indicating occasional failure to capture entire tokens, possibly due to coarticulation or rapid speech. Insertions (I) remain comparatively low across categories. Overall, the pattern suggests that the primary limitation of the Transformer model on this dataset lies in fine-grained lexical discrimination under phonetic and morphological variability, which is characteristic of agglutinative languages like Kannada. The high number of `x' errors indicate that the AI4Bharat PoS tagger has failed to classify the PoS of many of the words in the dataset.
\begin{table}[h]
\centering
\footnotesize
\setlength{\tabcolsep}{4pt}

\begin{tabular}{|c|c|r|r|r|c|}
\hline
\textbf{POS} & \textbf{Count} & \textbf{D} & \textbf{S} & \textbf{I} & \textbf{Total (\%)} \\
\hline
num     & 65   & 1  & 8   & 0 & 9 (13.85) \\
part    & 210  & 2  & 23  & 3 & 28 (13.33) \\
cconj   & 681  & 4  & 67  & 8 & 79 (11.60) \\
pron    & 1347 & 10 & 121 & 6 & 137 (10.17) \\
det     & 1565 & 23 & 123 & 9 & 155 (9.90) \\
sconj   & 151  & 2  & 10  & 1 & 13 (8.61) \\
adp     & 503  & 5  & 29  & 3 & 37 (7.36) \\
aux     & 700  & 8  & 23  & 5 & 36 (5.14) \\
adv     & 845  & 5  & 22  & 0 & 27 (3.20) \\
verb    & 2240 & 1  & 12  & 2 & 15 (0.67) \\
noun    & 2082 & 3  & 10  & 1 & 14 (0.67) \\
propn   & 180  & 1  & 0   & 0 & 1 (0.56) \\
adj     & 598  & 0  & 1   & 0 & 1 (0.17) \\
\hline
\end{tabular}
\caption{PoS-wise error statistics for Greek.}
\label{tab:greek_all_pos}
\end{table}

\textbf{Greek.} Table \ref{tab:greek_all_pos} demonstrates comparatively lower error percentages across all PoS categories. Although substitution errors remain the dominant error type, their magnitudes are much smaller than those observed for Hindi and Kannada. Categories such as \textit{num}, \textit{part}, \textit{cconj}, and \textit{pron} exhibit the highest total error percentages, namely \(13.85\%\), \(13.33\%\), \(11.60\%\), and \(10.17\%\), respectively. In contrast, major lexical categories such as \textit{verb} and \textit{noun} show very low total error percentages of only \(0.67\%\). Similarly, \textit{adj}, \textit{propn}, and \textit{punct} exhibit near-zero error rates. These results suggest that the Greek ASR system achieves relatively stable recognition performance across most grammatical categories, with errors concentrated mainly in smaller functional categories and low-frequency tokens.
\begin{table}[h]
\centering
\footnotesize
\setlength{\tabcolsep}{4pt}
\begin{tabular}{|c|c|r|r|r|c|}
\hline
\textbf{POS} & \textbf{Count} & \textbf{D} & \textbf{S} & \textbf{I} & \textbf{Total (\%)} \\
\hline
noun   & 8709 & 50 & 610 & 22 & 682 (7.83) \\
verb   & 5507 & 15 & 230 & 16 & 261 (4.74) \\
propn  & 8171 & 11 & 144 & 63 & 218 (2.67) \\
pron   & 5587 & 18 & 107 & 38 & 163 (2.92) \\
det    & 5561 & 23 & 77  & 56 & 156 (2.81) \\
adj    & 2567 & 12 & 116 & 11 & 139 (5.41) \\
adp    & 6134 & 8  & 83  & 31 & 122 (1.99) \\
aux    & 3364 & 12 & 58  & 8  & 78 (2.32) \\
adv    & 1928 & 6  & 51  & 8  & 65 (3.37) \\
cconj  & 2356 & 8  & 57  & 13 & 78 (3.31) \\
sconj  & 1139 & 2  & 16  & 5  & 23 (2.02) \\
part   & 1145 & 4  & 10  & 3  & 17 (1.48) \\
num    & 440  & 1  & 9   & 1  & 11 (2.50) \\
intj   & 212  & 0  & 7   & 3  & 10 (4.72) \\
x      & 15   & 3  & 6   & 0  & 9 (60.00) \\
\hline
\end{tabular}
\caption{PoS-wise error statistics for LibriSpeech test-clean.}
\label{tab:librispeech_test_clean_all_pos}
\end{table}

\textbf{LibriSpeech test-clean.} Table \ref{tab:librispeech_test_clean_all_pos} presents the PoS-wise ASR error statistics for the LibriSpeech test-clean subset. Among all PoS categories, \textit{noun} exhibits the highest total number of errors with 682 errors (\(7.83\%\)), primarily dominated by substitution errors (610). Although the absolute number of errors is large for nouns, the relative error percentage remains moderate due to the high occurrence count of nouns in the corpus. Verbs and adjectives also show comparatively higher error rates, with total error percentages of \(4.74\%\) and \(5.41\%\), respectively. Proper nouns (\textit{propn}) have a lower relative error rate of \(2.67\%\), indicating that the model performs comparatively well on named entities in clean acoustic conditions. Functional categories such as \textit{adp}, \textit{aux}, \textit{part}, and \textit{sconj} exhibit relatively low error percentages, suggesting that frequent syntactic tokens are recognized more reliably. The \textit{x} category shows the highest relative error percentage (\(60.00\%\)); however, this category contains only 15 tokens and therefore does not significantly affect overall WER. Overall, substitution errors dominate across nearly all PoS categories, indicating that acoustic confusions and lexical replacements are more common than deletions or insertions in clean speech conditions.
\begin{table}[h]
\centering
\footnotesize
\setlength{\tabcolsep}{4pt}
\begin{tabular}{|c|c|r|r|r|c|}
\hline
\textbf{POS} & \textbf{Count} & \textbf{D} & \textbf{S} & \textbf{I} & \textbf{Total (\%)} \\
\hline
noun   & 7920 & 71 & 1206 & 65 & 1342 (16.94) \\
verb   & 6275 & 37 & 549  & 50 & 636 (10.14) \\
propn  & 8211 & 18 & 384  & 173 & 575 (7.00) \\
pron   & 6209 & 56 & 281  & 76 & 413 (6.65) \\
det    & 5071 & 60 & 219  & 70 & 349 (6.88) \\
adj    & 2169 & 21 & 286  & 24 & 331 (15.26) \\
adp    & 5555 & 36 & 212  & 66 & 314 (5.65) \\
aux    & 3511 & 21 & 161  & 33 & 215 (6.12) \\
adv    & 2026 & 9  & 143  & 14 & 166 (8.19) \\
cconj  & 2298 & 19 & 89   & 35 & 143 (6.22) \\
sconj  & 1281 & 4  & 51   & 19 & 74 (5.78) \\
part   & 1283 & 10 & 34   & 10 & 54 (4.21) \\
num    & 512  & 3  & 17   & 6  & 26 (5.08) \\
intj   & 291  & 1  & 11   & 3  & 15 (5.15) \\
x      & 20   & 0  & 12   & 0  & 12 (60.00) \\
\hline
\end{tabular}
\caption{PoS-wise error statistics for LibriSpeech test-other.}
\label{tab:librispeech_test_other_all_pos}
\end{table}
\begin{table*}[h]
    \centering
    \begin{tabular}{|p{0.9cm}|p{1.0cm}|c|c|c|c|c|c|c|c|c|}
    \hline
    \multirow{2}{*}{\textbf{POS}} & \multirow{2}{*}{\textbf{count}} & \multicolumn{3}{c|}{\textbf{Transformer}} & \multicolumn{3}{c|}{\textbf{LSTM}} & \multicolumn{3}{c|}{\textbf{Conformer}}\\
    \cline{3-11}
    & & S & D & I & S & D & I & S & D & I\\
    \hline
    noun & 18905 & \cellcolor{rank1}15.4 & \cellcolor{rank1}2.38 & \cellcolor{rank1}3.2 & \cellcolor{rank1}16.61 & \cellcolor{rank1}2.68 & \cellcolor{rank1}2.85 & \cellcolor{rank1}17.7 & \cellcolor{rank1}61.15 & \cellcolor{rank1}0.16\\
    \hline
    propn & 15053 & \cellcolor{rank2}17.47 & \cellcolor{rank2}2.87 & \cellcolor{rank2}3.7 & \cellcolor{rank2}17.39 & \cellcolor{rank2}3.2 & \cellcolor{rank2}3.34 & \cellcolor{rank2}16.73 & \cellcolor{rank2}63.52 & \cellcolor{rank2}4.3\\
    \hline
    verb & 10365 & \cellcolor{rank3}13.73 & \cellcolor{rank3}1.76 & \cellcolor{rank3}2.6 & \cellcolor{rank3}13.62 & \cellcolor{rank3}2.0 & \cellcolor{rank3}2.25 & \cellcolor{rank3}29.76 & \cellcolor{rank3}46.99 & \cellcolor{rank3}0.0001\\
    \hline
    pron & 4816 & \cellcolor{rank4}10.55 & \cellcolor{rank4}3.18 & \cellcolor{rank4}4.17 & \cellcolor{rank5}10.20 & \cellcolor{rank5}3.11 & \cellcolor{rank5}3.55 & \cellcolor{rank6}14.04 & \cellcolor{rank6}72.2 & \cellcolor{rank6} 0\\
    \hline
    adj & 5316 & \cellcolor{rank5}10.57 & \cellcolor{rank5}2.48 & \cellcolor{rank5}3.07 & \cellcolor{rank4}11.5 &\cellcolor{rank4} 2.56 &\cellcolor{rank4}2.78 & \cellcolor{rank4}13.76 & \cellcolor{rank4}71.01 & \cellcolor{rank4} 0\\
    \hline
    adv & 5245 & \cellcolor{rank6}10.73 & \cellcolor{rank6}1.94 & \cellcolor{rank6}3.64 & \cellcolor{rank6}10.73 & \cellcolor{rank6}1.94 & \cellcolor{rank6}2.42 & \cellcolor{rank5}15.46 & \cellcolor{rank5}67.5 & \cellcolor{rank5}0.05\\
    \hline
    \end{tabular}
    \vspace{0.5em}
    \begin{tabular}{l l l l l l l}
    \cellcolor{rank1}\phantom{x} Rank 1 &
    \cellcolor{rank2}\phantom{x} Rank 2 &
    \cellcolor{rank3}\phantom{x} Rank 3 &
    \cellcolor{rank4}\phantom{x} Rank 4 &
    \cellcolor{rank5}\phantom{x} Rank 5 &
    \cellcolor{rank6}\phantom{x} Rank 6 &
    \end{tabular}
    \caption{Error distributions of different models for SPRING-INX Tamil dataset.}
    \label{table:err_analysis_tamil_diff_models}
\end{table*}

\textbf{LibriSpeech test-other.} Table \ref{tab:librispeech_test_other_all_pos} presents the PoS-wise ASR error statistics for the LibriSpeech test-other subset, which is acoustically more challenging than test-clean. Compared to test-clean, almost all PoS categories show noticeably higher error percentages. Nouns again contribute the largest number of total errors with 1342 errors (\(16.94\%\)), dominated by substitution errors (1206). Verbs also exhibit a substantial increase in errors, reaching 636 total errors (\(10.14\%\)). Adjectives show a relatively high error percentage of \(15.26\%\), indicating increased difficulty in recognizing descriptive lexical items under challenging acoustic conditions. Proper nouns (\textit{propn}) have a total error percentage of \(7.00\%\), which is considerably higher than in test-clean, suggesting greater sensitivity of named entities to acoustic degradation. Function words such as \textit{det}, \textit{adp}, \textit{aux}, and \textit{cconj} also experience increased substitution and insertion errors compared to test-clean. Similar to the clean subset, substitution errors remain the dominant error type across nearly all PoS categories. 

Overall, the results indicate that acoustically difficult speech conditions disproportionately affect content-bearing PoS categories such as nouns, verbs, and adjectives.

\begin{table*}[t]
\centering
\label{tab:combined_pos_weights}
\footnotesize
\setlength{\tabcolsep}{4pt}
\begin{tabular}{|lc|lc|lc|}
\hline
\multicolumn{2}{|c|}{\textbf{Tamil}} &
\multicolumn{2}{|c|}{\textbf{Arabic}} &
\multicolumn{2}{|c|}{\textbf{Russian}} \\
\hline
\textbf{PoS} & \textbf{Weight} &
\textbf{PoS} & \textbf{Weight} &
\textbf{PoS} & \textbf{Weight} \\
\hline
NOUN   & 4.0 & NOUN           & 8.0 & NOUN    & 6.0 \\
PROPN  & 4.0 & NOUN\_PROP     & 2.0 & VERB    & 6.0 \\
VERB   & 2.0 & VERB           & 8.0 & ADJ     & 1.0 \\
AUX    & 1.0 & VERB\_PSEUDO   & 5.0 & PROPN   & 1.0 \\
ADJ    & 1.5 & ADJ            & 2.5 & ADP     & 2.0 \\
ADV    & 1.5 & ADV            & 1.5 & PRON    & 2.0 \\
PRON   & 2.0 & PRON           & 1.5 & CCONJ   & 1.0 \\
ADP    & 1.0 & PRON\_DEM      & 1.5 & SCONJ   & 1.0 \\
DET    & 1.0 & PRON\_EXCLAM   & 1.5 & ADV     & 1.5 \\
CONJ   & 1.0 & PRON\_INTERROG & 1.5 & AUX     & 1.0 \\
CCONJ  & 1.0 & PREP           & 2.0 & PART    & 1.0 \\
NUM    & 1.0 & CONJ           & 1.5 & NUM     & 1.0 \\
PART   & 1.0 & PART           & 1.0 & DET     & 1.0 \\
SYM    & 1.0 & PART\_NEG      & 1.0 & INTJ    & 1.0 \\
INTJ   & 1.0 & PART\_FOCUS    & 1.0 & X       & 1.0 \\
X      & 1.0 & PART\_INTERROG & 1.0 & SYM     & 1.0 \\
        &     & PART\_FUT      & 1.0 & UNKNOWN & 1.0 \\
        &     & NUM            & 1.0 &         &     \\
        &     & ABBREV         & 1.0 &         &     \\
        &     & FOREIGN        & 1.0 &         &     \\
        &     & INTERJ         & 1.0 &         &     \\
\hline
\end{tabular}
\caption{PoS-specific weights used for decoder-side PoS-aware attention for Tamil, Arabic, and Russian.}
\label{tab:pos_weights}
\end{table*}
\section{Error analysis of different models for Tamil dataset}\label{ap:error_analysis_tamil}
For comparing PoS error distribution across models for Tamil ASR, we consider the IndicConformer model\footnote{\href{https://huggingface.co/ai4bharat/indicconformer_stt_ta_hybrid_ctc_rnnt_large}{IndicConformer Tamil ASR model}} comprising of 17 Conformer blocks of model dimension 512. We further use an LSTM model of standard configuration\footnote{\href{https://github.com/espnet/espnet/blob/effbd689ceaa2ee1131369c2c64acc30e6016706/egs2/aishell/asr1/conf/train_asr_rnn.yaml\#L4}{LSTM Tamil ASR model}}.
A comparison across Transformer, LSTM, and Conformer models in Table \ref{table:err_analysis_tamil_diff_models} reveals consistent patterns in error distribution, while also highlighting architecture-specific weaknesses. Across all three models, nouns and proper nouns remain the most error-prone categories, indicating that lexical variability and agglutinative morphology are primary sources of difficulty in Tamil ASR. Both Transformer and LSTM exhibit similar behavior, with dominating substitution errors and relatively moderate deletion and insertion errors. This suggests that Transformer and LSTM are reasonably effective at capturing token boundaries but struggle with lexical precision. In contrast, Conformer displays a different error profile, characterized by an unusually high deletion errors, especially for nouns, proper nouns, and verbs. This indicates a tendency to omit tokens entirely rather than substitute them, possibly due to alignment issues in longer sequences. Interestingly, significant reduction in insertion errors (often near zero) by Conformer, comes at the cost of aggressive deletions. Pronouns, adjectives, and adverbs show relatively lower error counts across all models, likely due to their higher frequency and more constrained usage patterns, though Conformer exhibits high deletions even in these categories. Overall, while Transformer and LSTM maintain a balanced error distribution dominated by substitutions, Conformer introduces a skewed trade-off between deletions and insertions, suggesting that improvements in acoustic modeling do not necessarily translate to better lexical retention in morphologically rich languages like Tamil.


\section{PoS weights and model details} \label{sec:appendix_poswts}
For incorporating PoS weights, we use a standard Transformer architecture with a 12-layer encoder and 6-layer decoder\footnote{\href{https://github.com/espnet/espnet/blob/master/egs2/librispeech/asr1/conf/tuning/train\_asr\_transformer.yaml}{Transformer 12 layer encoder and 6 layer decoder}}. The encoder has a model dimension of 512 with 8 attention heads, with the same configuration for the decoder. 
Training is performed using the Adam optimizer with a learning rate of 0.002 and a warm-up scheduler with 25k warm-up steps. We use gradient accumulation of 4 and a batch size of 16M bins. 
SpecAugment is applied with time warping, frequency masking, and time masking to improve robustness. For all experiments, $\lambda_{\text{CTC}}$ is set as $0.3$.
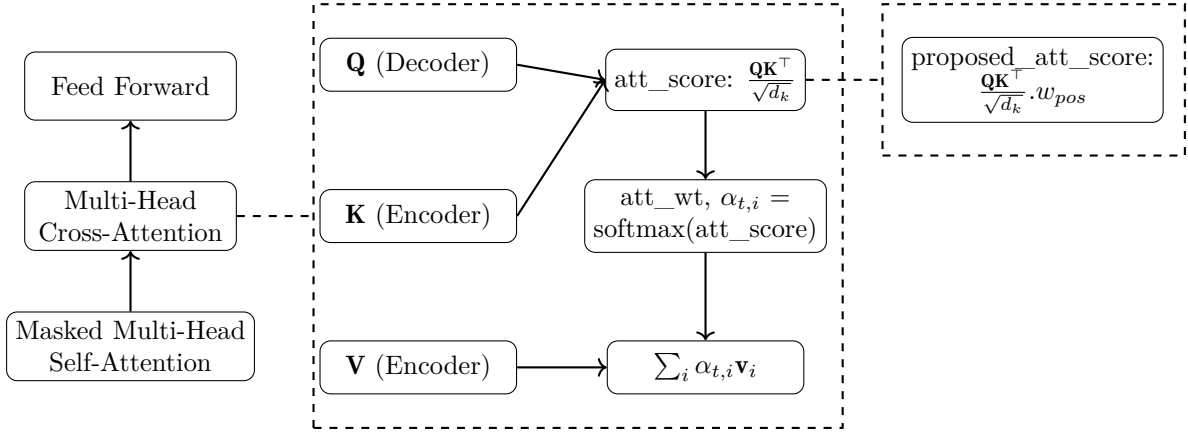
\begin{figure*}[h]
\centering
\begin{tikzpicture}[
    box/.style={draw, rounded corners, minimum width=2.8cm, minimum height=0.9cm, align=center},
    smallbox/.style={draw, rounded corners, minimum width=2.6cm, minimum height=0.7cm, align=center},
    arrow/.style={->, thick},
    zoom/.style={draw, dashed, thick}
]
\node[box] (msa) {Masked Multi-Head\\Self-Attention};
\node[box, above=0.8cm of msa] (ca) {Multi-Head\\ Cross-Attention};
\node[box, above=0.8cm of ca] (ffn) {Feed Forward};

\draw[arrow] (msa) -- (ca);
\draw[arrow] (ca) -- (ffn);
\node[zoom, right=1.0cm of ca, minimum width=7cm, minimum height=5.6cm] (zoomBox) {};
\draw[dashed, thick] (ca.east) -- (zoomBox.west);
\node[smallbox] (q) at ($(zoomBox.west)+(1.4,2.0)$) {$\mathbf{Q}$ (Decoder)};
\node[smallbox] (k) at ($(zoomBox.west)+(1.4,0.0)$) {$\mathbf{K}$ (Encoder)};
\node[smallbox] (v) at ($(zoomBox.west)+(1.4,-2.0)$) {$\mathbf{V}$ (Encoder)};
\node[smallbox] (score) at ($(zoomBox.west)+(5.2,1.8)$)
{att\_score: $\frac{\mathbf{Q}\mathbf{K}^\top}{\sqrt{d_k}}$};
\node[smallbox] (softmax) at ($(zoomBox.west)+(5.2,0.0)$)
{att\_wt, $\alpha_{t,i}$ =\\ softmax(att\_score) };
\node[smallbox] (output) at ($(zoomBox.west)+(5.2,-2.0)$)
{$\sum_i \alpha_{t,i}\mathbf{v}_i$};
\draw[arrow] (q.east) -- (score.west);
\draw[arrow] (k.east) -- (score.west);
\draw[arrow] (score) -- (softmax);
\draw[arrow] (softmax) -- (output);
\draw[arrow] (v.east) -- (output.west);
\node[zoom, right=1.0cm of score, minimum width=4cm, minimum height=2cm] (zoomBox2) {};
\draw[dashed, thick] (score.east) -- (zoomBox2.west);
\node[smallbox] (our_score) at ($(zoomBox2.west)+(2.0,0.0)$) {proposed\_att\_score:\\ $\frac{\mathbf{Q}\mathbf{K}^\top}{\sqrt{d_k}}.w_{pos}$};
\end{tikzpicture}
\caption{{Col-1: Default decoder layers in Transformer; Col-2: cross-attention with attention score computation; Col-3: modified attention score using scalar PoS weights, $w_{pos}$}}
\label{fig:decoder_cross_attention_zoom}
\end{figure*}
\begin{table}[h!]
\centering
\footnotesize
\setlength{\tabcolsep}{10pt}
\begin{tabular}{|l|c|c|c|}
\hline
\textbf{Dataset} & \textbf{Train} & \textbf{Validation} & \textbf{Test} \\
\hline
Tamil & 56,112 & 16,699 & 4,884 \\
\hline
Arabic & 28,865 & 10,229 & 10,508 \\
\hline
Russian & 26,920 & 10,282 & 10,283 \\
\hline
\end{tabular}
\caption{Number of utterances in the train, validation, and test splits for the evaluated datasets.}
\label{tab:dataset_split_statistics}
\end{table}
The weights chosen for SPRING INX-Tamil, Commonvoice Arabic and Russian datasets is given in Table \ref{tab:pos_weights}. For PoS-aware training, these weights are computed token-wise for each test dataset in prior and stored as a JSON file, which is then used by the transformer. All experiments were conducted on a A100 GPU with 80 GB RAM. The training time is $\sim 40$ hours, $\sim 17$ hours, and $\sim 58$ hours for 100 epochs for Tamil, Arabic, and Russian dataset respectively. The train, validation and test splits for the datasets used in training are provided in Table \ref{tab:dataset_split_statistics}.

\section{PoS-wise distribution after Decoder PoS fusion}\label{ap:decoder_pos}
An illustration of the Decoder PoS fusion in the Transformer encoder-decoder architecture is provided in Fig. \ref{fig:decoder_cross_attention_zoom}.
\begin{table*}[h]
\centering
\footnotesize
\setlength{\tabcolsep}{6pt}
\begin{tabular}{@{}|p{1.0cm}|p{1.0cm}|rrrr|rrrr|@{}}
\hline
\multirow{2}{*}{\textbf{PoS}} &
\multirow{2}{*}{\textbf{Count}} &
\multicolumn{4}{c|}{\textbf{Without PoS}} &
\multicolumn{4}{c|}{\textbf{Decoder with PoS}} \\
\cline{3-10}
& & D & S & I & Total (\%) & D & S & I & Total (\%) \\
\hline
noun  & 18905 & 451 & 2911 & 605 & 3967 (20.98) & 470 & \textbf{2807} & \textbf{546} & 3823 (20.22) \\
propn & 15053 & 433 & 2630 & 557 & 3620 (24.05) & 474 & \textbf{2419} & \textbf{532} & 3425 (22.75) \\
verb  & 10365 & 182 & 1423 & 270 & 1875 (18.09) & 205 & \textbf{1351} & \textbf{240} & 1796 (17.33) \\
pron  & 4816  & 153 & 508  & 201 & 862 (17.90)  & 165 & \textbf{494} & \textbf{161} & 820 (17.03)  \\
adj   & 5316  & 132 & 562  & 163 & 857 (16.12)  & \textbf{127} & 562 & \textbf{143} & 832 (15.65)  \\
adv   & 5245  & 102 & 563  & 191 & 856 (16.32)  & 117 & \textbf{524} & \textbf{151} & 792 (15.10)  \\
part  & 3863  & 87  & 338  & 103 & 528 (13.67)  & 98  & 338 & 104 & 540 (13.98)  \\
num   & 1748  & 74  & 366  & 66  & 506 (28.95)  & \textbf{70} & \textbf{351} & \textbf{63} & 484 (27.69)   \\
aux   & 2789  & 37  & 284  & 73  & 394 (14.13)  & 37  & \textbf{257} & \textbf{64} & 358 (12.84)  \\
adp   & 1432  & 25  & 233  & 33  & 291 (20.32)  & 34  & \textbf{217} & \textbf{32} & 283 (19.76)  \\
det   & 2131  & 29  & 153  & 82  & 264 (12.39)  & 29  & 164 & \textbf{58} & 251 (11.78) \\
cconj & 188   & 1   & 3    & 2   & 6 (3.19)     & 1   & \textbf{2} & \textbf{1} & 4 (2.13) \\
\hline
\end{tabular}
\caption{PoS-wise error comparison for SPRING-INX Tamil dataset; without PoS refers to baseline; Decoder with PoS: PoS incorporated in Transformer decoder}
\label{tab:pos_full_comparison_tamil}
\end{table*}

\textbf{Tamil.}
\begin{figure*}[h]
\centering
\begin{tikzpicture}
\begin{axis}[
    ybar,
    bar width=6pt,
    width=0.8\textwidth,
    height=5cm,
    enlarge x limits=0.02,
    legend style={at={(0.5,1.15)}, anchor=south, legend columns=3},
    ylabel={Error (\%)},
    symbolic x coords={noun,propn,verb,pron,adj,adv,part,num,aux,adp,det,cconj},
    xtick=data,
    x tick label style={rotate=90, anchor=east},
    ymin=0,
    ymax=35,
    grid=both,
]
\addplot[fill=orange!50] coordinates {
(noun,20.98) (propn,24.05) (verb,18.09) (pron,17.90)
(adj,16.12) (adv,16.32) (part,13.67)
(num,28.95) (aux,14.13) (adp,20.32) (det,12.39) (cconj,3.19)
};
\addplot[fill=green!50] coordinates {
(noun,20.22) (propn,22.75) (verb,17.33) (pron,17.03)
(adj,15.65) (adv,15.10) (part,13.98)
(num,27.69) (aux,12.84) (adp,19.76) (det,11.78) (cconj,2.13)
};
\legend{Without PoS, Decoder with PoS}
\end{axis}
\end{tikzpicture}
\caption{PoS-wise error comparison for SPRING-INX Tamil dataset;x-axis: PoS tags; y-axis: corresponding error.}
\label{fig:pos_bar_chart_tamil}
\end{figure*}
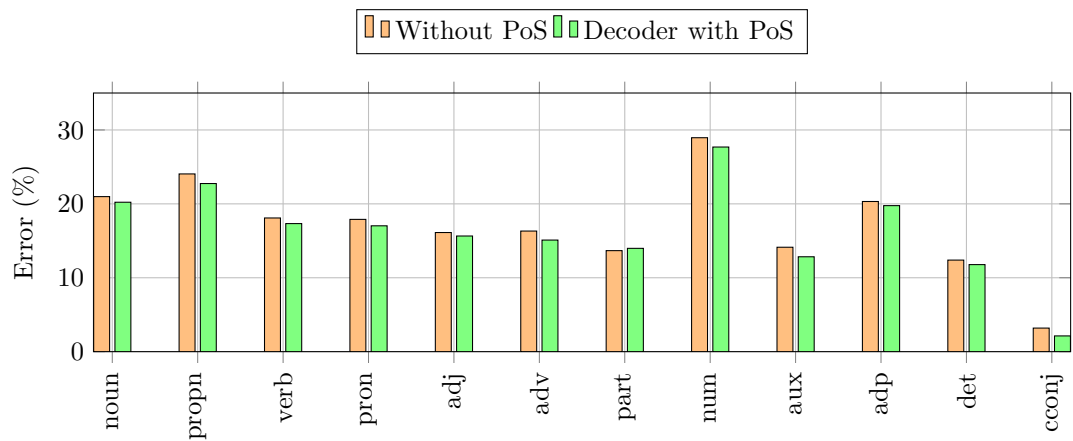

Table \ref{tab:pos_full_comparison_tamil} presents the PoS-wise error analysis for the SPRING-INX Tamil dataset. Categories with a reduction in the number of errors are highlighted in bold. Across most PoS categories, incorporating PoS information in the decoder reduces substitution errors, which remain the dominant error type. Significant improvements are observed for content-bearing categories such as \textit{noun}, \textit{propn}, \textit{verb}, \textit{adv}, and \textit{punct}. For instance, substitution errors for \textit{propn} decrease from 2630 to 2419, corresponding to a relative reduction of approximately $8.02\%$. Similarly, substitution errors for \textit{verb} reduce from 1423 to 1351 ($5.06\%$ reduction), while \textit{noun} substitutions decrease from 2911 to 2807 ($3.57\%$ reduction). Categories such as \textit{adv}, \textit{aux}, \textit{num}, and \textit{punct} also exhibit consistent reductions in both insertion and substitution errors. Although a few categories such as \textit{part} show a slight increase in deletions, the overall total error percentage decreases for most major PoS groups. These results suggest that decoder-level PoS conditioning helps the model better capture syntactic and lexical structure, thereby improving recognition accuracy and contributing to the overall WER reduction from 23.3 to 22.0. For a graphical comparison of reduction in PoS-wise errors, refer to Fig. \ref{fig:pos_bar_chart_tamil}.
\begin{table}[h]
\centering
\footnotesize
\begin{tabular}{|l|p{5cm}|}
\hline
\textbf{Tag} & \textbf{Description} \\
\hline
\textit{noun\_prop} & Proper noun (names of people, places, organizations) \\
\hline
\textit{noun\_quant} & Quantifier noun indicating quantity or amount \\
\hline
\textit{verb\_pseudo} & Pseudo-verbs or quasi-verbal particles in Arabic grammar \\
\hline
\textit{part\_det} & Determiner particle \\
\hline
\textit{part\_neg} & Negation particle \\
\hline
\textit{part\_voc} & Vocative particle used for addressing \\
\hline
\textit{part\_verb} & Verbal particle associated with tense/aspect constructions \\
\hline
\textit{part\_focus} & Focus or emphasis particle \\
\hline
\textit{part\_interrog} & Interrogative particle used in questions \\
\hline
\textit{pron\_dem} & Demonstrative pronoun \\
\hline
\textit{pron\_rel} & Relative pronoun \\
\hline
\textit{pron\_interrog} & Interrogative pronoun \\
\hline
\textit{conj\_sub} & Subordinating conjunction \\
\hline
\textit{adv\_rel} & Relative adverb \\
\hline
\textit{adv\_interrog} & Interrogative adverb \\
\hline
\textit{interj} & Interjection \\
\hline
\textit{foreign} & Foreign or non-Arabic word \\
\hline
\textit{abbrev} & Abbreviation \\
\hline
\end{tabular}
\caption{Fine-grained non-UPoS tags used by the Arabic PoS tagger.}
\label{tab:arabic_nonupos}
\end{table}

\textbf{Arabic.}
The non-UPoS tags used by the Arabic PoS tagger correspond to fine-grained grammatical categories specific to Arabic morphology, described in Table \ref{tab:arabic_nonupos}.
\begin{table*}[h]
\centering
\footnotesize
\setlength{\tabcolsep}{5pt}
\begin{tabular}{@{}|p{2.1cm}|p{1.0cm}|rrrr|rrrr|@{}}
\hline
\multirow{2}{*}{\textbf{PoS}} &
\multirow{2}{*}{\textbf{Count}} &
\multicolumn{4}{c|}{\textbf{Baseline}} &
\multicolumn{4}{c|}{\textbf{Decoder with PoS}} \\
\cline{3-10}
& & D & S & I & Total (\%) & D & S & I & Total (\%) \\
\hline

noun &
16676 &
\textbf{274} & 6197 & 189 & 6660 (39.94)
& 287 & \textbf{6056} & \textbf{176} & \textbf{6519 (39.09)} \\

verb &
10741 &
\textbf{244} & 4690 & 175 & 5109 (47.57)
& 264 & \textbf{4679} & \textbf{164} & \textbf{5107 (47.55)} \\

noun\_prop &
8919 &
\textbf{132} & 2546 & 103 & 2781 (31.18)
& 147 & \textbf{2491} & \textbf{102} & \textbf{2740 (30.72)} \\

prep &
5378 &
\textbf{105} & 1817 & \textbf{58} & 1980 (36.82)
& 126 & \textbf{1769} & 67 & \textbf{1962 (36.48)} \\

adj &
2380 &
\textbf{28} & \textbf{747} & \textbf{17} & \textbf{792 (33.28)}
& 31 & 757 & 20 & 808 (33.95) \\

verb\_pseudo &
134 &
13 & 163 & \textbf{7} & 183 (136.57)
& \textbf{9} & \textbf{159} & 8 & \textbf{176 (131.34)} \\

part\_neg &
1229 &
\textbf{39} & 586 & 41 & 666 (54.19)
& 55 & \textbf{571} & \textbf{31} & \textbf{657 (53.46)} \\

pron\_rel &
1177 &
49 & \textbf{534} & 57 & \textbf{640 (54.38)}
& \textbf{44} & 543 & \textbf{56} & 643 (54.63) \\


conj\_sub &
1680 &
\textbf{21} & \textbf{466} & 20 & \textbf{507 (30.18)}
& 27 & 468 & 20 & 515 (30.65) \\

pron &
903 &
32 & 400 & \textbf{21} & 453 (50.17)
& 33 & \textbf{388} & 26 & \textbf{447 (49.50)} \\

conj &
903 &
\textbf{36} & 370 & 34 & \textbf{440 (48.73)}
& 47 & \textbf{366} & \textbf{30} & 443 (49.06) \\

part\_voc &
479 &
\textbf{49} & 298 & 32 & 379 (79.12)
& 68 & \textbf{267} & \textbf{31} & \textbf{366 (76.41)} \\

part\_interrog &
355 &
38 & \textbf{154} & \textbf{52} & \textbf{244 (68.73)}
& \textbf{34} & 162 & 63 & 259 (72.96) \\

part\_verb &
358 &
\textbf{14} & \textbf{187} & 13 & \textbf{214 (59.78)}
& 21 & 192 & \textbf{11} & 224 (62.57) \\

abbrev &
51 &
\textbf{19} & \textbf{164} & \textbf{12} & \textbf{195 (382.35)}
& 24 & 169 & 13 & 206 (403.92) \\

adv &
336 &
19 & 123 & 7 & 149 (44.35)
& \textbf{18} & \textbf{113} & \textbf{3} & \textbf{134 (39.88)} \\

pron\_interrog &
110 &
14 & \textbf{64} & \textbf{15} & \textbf{93 (84.55)}
& \textbf{13} & 67 & 20 & 100 (90.91) \\

part &
201 &
2 & \textbf{66} & 1 & \textbf{69 (34.33)}
& \textbf{1} & 78 & \textbf{0} & 79 (39.30) \\

adv\_rel &
79 &
9 & \textbf{41} & \textbf{11} & \textbf{61 (77.22)}
& \textbf{8} & 55 & 21 & 84 (106.33) \\

adv\_interrog &
117 &
9 & 39 & \textbf{7} & \textbf{55 (47.01)}
& \textbf{7} & 39 & 10 & 56 (47.86) \\

part\_focus &
39 &
\textbf{2} & 33 & 3 & 38 (97.44)
& 5 & \textbf{29} & \textbf{1} & \textbf{35 (89.74)} \\

foreign &
31 &
2 & 20 & 0 & 22 (70.97)
& 2 & \textbf{19} & 0 & \textbf{21 (67.74)} \\

interj &
20 &
2 & \textbf{11} & 5 & 18 (90.00)
& \textbf{1} & 13 & \textbf{2} & \textbf{16 (80.00)} \\

part\_fut &
17 &
\textbf{1} & \textbf{8} & 1 & \textbf{10 (58.82)}
& 2 & 9 & \textbf{0} & 11 (64.71) \\

noun\_quant &
111 &
0 & \textbf{7} & \textbf{0} & \textbf{7 (6.31)}
& 0 & 8 & 1 & 9 (8.11) \\

part\_det &
18 &
\textbf{0} & \textbf{1} & \textbf{0} & \textbf{1 (5.56)}
& -- & -- & -- & -- \\
\hline
\end{tabular}
\caption{PoS-wise error comparison for the Arabic dataset. Baseline refers to the standard Transformer ASR model, while Decoder with PoS denotes Transformer decoder training augmented with PoS information. Boldface indicates lower error counts between the two systems for each metric.}
\label{tab:arabic_pos_full_comparison}
\end{table*}
Table \ref{tab:arabic_pos_full_comparison} presents the PoS-wise error analysis for the Arabic dataset, where lower error counts between the two systems are highlighted in bold. Similar to the Tamil experiments, substitution errors dominate across most PoS categories. Incorporating PoS information into the decoder reduces errors for several major lexical categories such as \textit{noun}, \textit{verb}, \textit{noun\_prop}, and \textit{prep}, which contribute significantly to the overall WER. In particular, total errors for \textit{noun} decrease from 6660 to 6519 ($2.12\%$ reduction), \textit{noun\_prop} decreases from 2781 to 2740 ($1.47\%$ reduction), and \textit{prep} decreases from 1980 to 1962 ($0.91\%$ reduction). Function-word categories such as \textit{adv}, \textit{part\_focus}, and \textit{foreign} also show improvements, mainly through reduced substitution errors, indicating that PoS-aware decoding helps the model better capture syntactic and lexical context during prediction. Although a few categories such as \textit{adj}, \textit{conj\_sub}, and \textit{part\_interrog} exhibit slight increases in errors, the dominant content-bearing categories show overall improvement, contributing to the WER reduction from 41.6 to 41.4.

\begin{table*}[h]
\centering
\footnotesize
\setlength{\tabcolsep}{5pt}
\begin{tabular}{@{}|p{1.8cm}|p{1.0cm}|rrrr|rrrr|@{}}
\hline
\multirow{2}{*}{\textbf{PoS}} &
\multirow{2}{*}{\textbf{Count}} &
\multicolumn{4}{c|}{\textbf{Baseline}} &
\multicolumn{4}{c|}{\textbf{Decoder with PoS}} \\
\cline{3-10}
& & D & S & I & Total (\%) & D & S & I & Total (\%) \\
\hline

propn &
60150 &
416 & 3007 & \textbf{331} & 3754 (6.24)
& \textbf{361} & \textbf{2886} & 360 & \textbf{3607 (6.00)} \\

noun &
11946 &
\textbf{84} & 557 & \textbf{69} & 710 (5.94)
& 89 & \textbf{511} & 73 & \textbf{673 (5.63)} \\

adp &
6063 &
239 & \textbf{206} & \textbf{112} & 557 (9.19)
& \textbf{206} & 209 & 120 & \textbf{535 (8.82)} \\

cconj &
3141 &
150 & \textbf{120} & 71 & 341 (10.86)
& \textbf{142} & 126 & \textbf{55} & \textbf{323 (10.28)} \\


x &
617 &
9 & \textbf{221} & \textbf{5} & \textbf{235 (38.09)}
& \textbf{5} & 232 & 6 & 243 (39.38) \\

part &
1430 &
31 & \textbf{44} & \textbf{23} & 98 (6.85)
& \textbf{16} & 45 & 29 & \textbf{90 (6.29)} \\

pron &
2015 &
29 & 48 & \textbf{13} & 90 (4.47)
& \textbf{24} & \textbf{40} & 15 & \textbf{79 (3.92)} \\

det &
595 &
\textbf{6} & 22 & 2 & 30 (5.04)
& 9 & \textbf{18} & \textbf{0} & \textbf{27 (4.54)} \\

verb &
572 &
6 & 16 & \textbf{3} & \textbf{25 (4.37)}
& \textbf{4} & 16 & 6 & 26 (4.55) \\

adj &
90 &
1 & \textbf{3} & \textbf{0} & \textbf{4 (4.44)}
& \textbf{0} & 8 & \textbf{0} & 8 (8.89) \\

adv &
240 &
2 & 5 & 1 & 8 (3.33)
& \textbf{1} & \textbf{4} & 1 & \textbf{6 (2.50)} \\

sconj &
72 &
\textbf{0} & \textbf{0} & 1 & \textbf{1 (1.39)}
& 1 & \textbf{0} & 1 & 2 (2.78) \\

num &
2 &
0 & 1 & 0 & 1 (50.00)
& 0 & 1 & 0 & 1 (50.00) \\
\hline
\end{tabular}
\caption{PoS-wise error comparison for the Russian dataset. Baseline refers to the standard Transformer ASR model, while Decoder with PoS denotes Transformer decoder training augmented with PoS information. Boldface indicates lower error counts between the two systems for each metric.}
\label{tab:russian_pos_full_comparison}
\end{table*}

\textbf{Russian.}
Table \ref{tab:russian_pos_full_comparison} presents the PoS-wise error analysis for the Russian dataset. Similar to the Tamil and Arabic experiments, substitution errors remain the dominant error type across most PoS categories. Incorporating PoS information into the decoder reduces total errors for several major lexical and functional categories, particularly \textit{propn}, \textit{noun}, \textit{adp}, \textit{cconj}, \textit{pron}, and \textit{part}. 

The largest improvement is observed for \textit{propn}, where total errors decrease from 3754 to 3607, corresponding to a relative reduction of approximately $3.92\%$. Likewise, \textit{noun} errors reduce from 710 to 673 ($5.21\%$ reduction), while \textit{adp} errors decrease from 557 to 535 ($3.95\%$ reduction). Categories such as \textit{pron} and \textit{part} also exhibit consistent reductions in deletion and substitution errors, leading to overall total error reductions of approximately $12.22\%$ and $8.16\%$, respectively. The \textit{det} category additionally benefits from reduced substitution and insertion errors, decreasing total errors from 30 to 27.

However, the gains are not uniform across all categories. Certain low-frequency PoS groups such as \textit{adj}, \textit{verb}, and \textit{sconj} show slight increases in total errors, mainly due to increased substitutions or insertions. Nevertheless, across the major content-bearing and syntactically important categories, PoS-aware decoder training generally reduces substitution and deletion errors, contributing to the overall WER improvement from 9.7 to 9.2.
\section{Statement on Generative AI Use}
Generative AI tools were used solely for grammar checking and language refinement.

\end{document}